\pdfoutput=1

\documentclass[11pt]{article}

\usepackage[final]{acl}

\usepackage{times}
\usepackage{latexsym}

\usepackage[T1]{fontenc}
\usepackage[greek,english]{babel}
\usepackage{kotex}

\usepackage[utf8]{inputenc}

\usepackage[nopatch=footnote]{microtype}

\usepackage{inconsolata}

\usepackage{graphicx}
\usepackage{linguex}

\usepackage{booktabs}

\usepackage{fontawesome}

\title{On the Acquisition of Shared Grammatical Representations \\ in Bilingual Language Models
}

\author{Catherine Arnett$^{a,b}$ ~~ Tyler A. Chang$^{c}$  ~~ James A. Michaelov$^{c,d, e}$~~  Benjamin K. Bergen$^{c}$\\
  $^{a}$Department of Linguistics, UCSD  ~~~$^{b}$EleutherAI \\
  $^{c}$Department of Cognitive Science, UCSD \\ 
  $^{d}$Department of Brain and Cognitive Sciences, MIT ~~~$^{e}$MIT Libraries CREOS\\
  \texttt{catherine@eleuther.ai, tachang@ucsd.edu, jamic@mit.edu, bkbergen@ucsd.edu}   }

\begin{document}
\maketitle

\begin{abstract}
Crosslingual transfer is crucial to contemporary language models' multilingual capabilities, but how it occurs is not well understood. We ask what happens to a monolingual language model when it begins to be trained on a second language. Specifically, we train small bilingual models for which we control the amount of data for each language and the order of language exposure.
To find evidence of shared multilingual representations, we turn to structural priming, a method used to study grammatical representations in humans. 
We first replicate previous crosslingual structural priming results and find that after controlling for training data quantity and language exposure, there are asymmetrical effects across language pairs and directions. We argue that this asymmetry may shape hypotheses about human structural priming effects. We also find that structural priming effects are less robust for less similar language pairs, highlighting potential limitations of crosslingual transfer learning and shared representations for typologically diverse languages.
\end{abstract}

\vspace{-0.6em}
\begin{center}
   \raisebox{-2.2pt}{\includegraphics[scale=0.09]{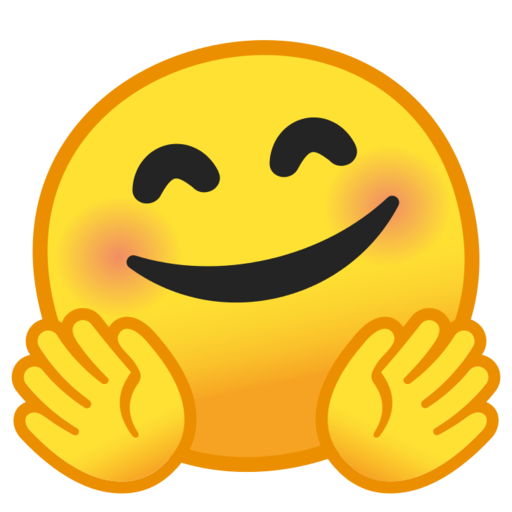}} \href{https://huggingface.co/collections/catherinearnett/b-gpt-66f4b80e8fa8e95491948556}{~\texttt{B-GPT models}} ~~~ \raisebox{-0.2ex}{\includegraphics[height=1em]{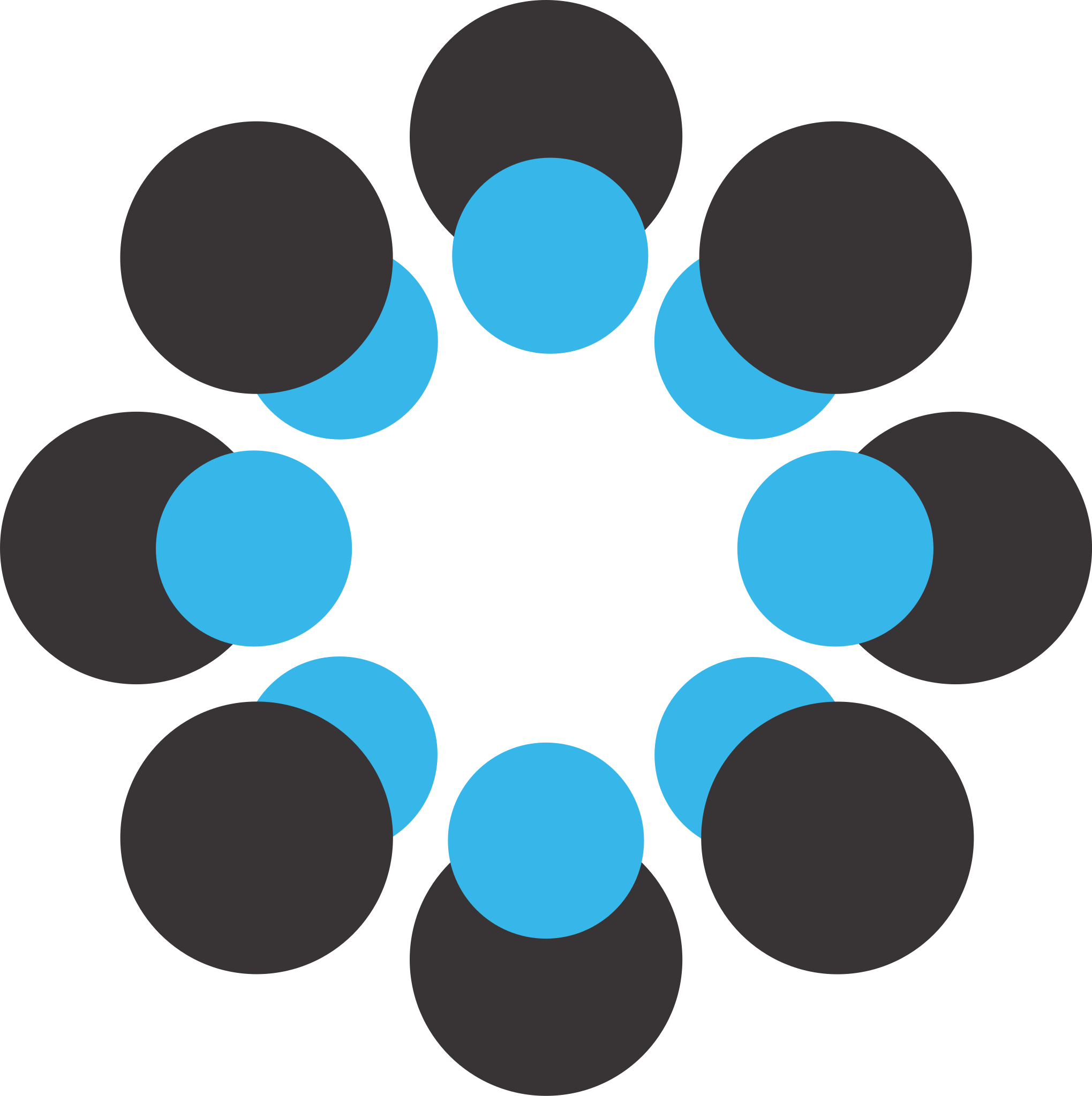}}
 \href{https://osf.io/5cw2e/}{\texttt{code and data}} 
\end{center}
\vspace{-0.5em}

\section{Introduction}

\begin{figure*}[ht!]
    \centering
    \vskip -1em
    \includegraphics[width=0.85\textwidth]{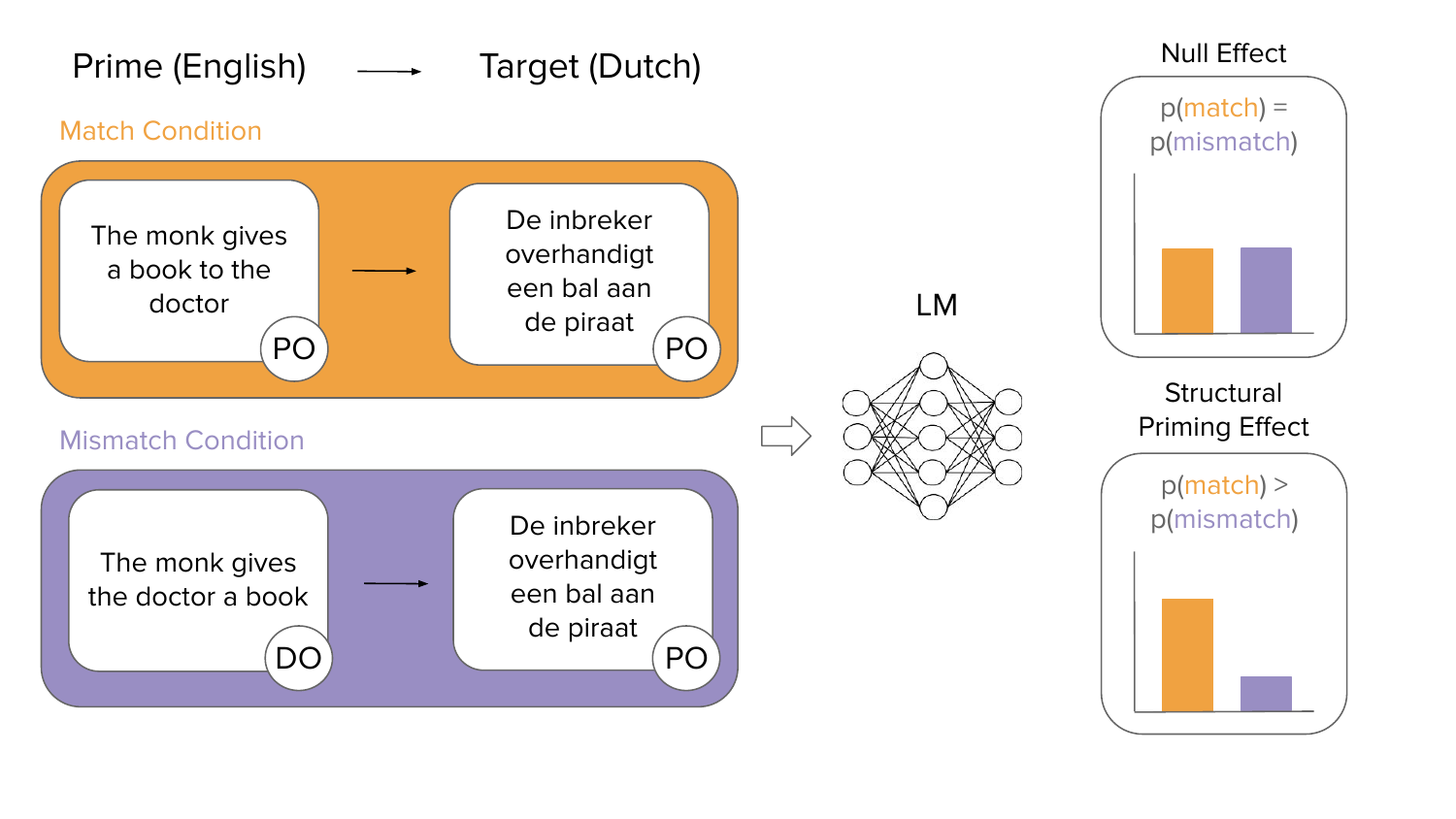}
    \vskip -1.5em
    \caption{Structural priming paradigm.}
    \label{fig:diagram}
    
\end{figure*}

Multilingual language models share representations across languages \citep{artetxe-etal-2020-cross, conneau-etal-2020-emerging}, which is thought to be crucial for crosslingual transfer abilities \citep{wu-dredze-2019-beto,Chi_Dong_Wei_Wang_Mao_Huang_2020,pmlr-v119-hu20b,winata-etal-2021-language,winata-etal-2022-cross}.
While there has been much evidence that successful crosslingual transfer can enable improvements in performance, there has not yet been extensive research about how models develop the shared representations that drive it. 

To study these shared representations, we use structural priming, a phenomenon in which a target sentence with a congruent preceding prime sentence type will have a higher likelihood than the same target sentence following an incongruent prime.
For example, we predict a language model would assign a higher probability to a prepositional object (PO) dative sentence (e.g. ``\textit{the chef gives a hat to the swimmer}'') following another PO sentence than following a double object (DO) dative sentence (e.g. ``\textit{the chef gives the swimmer a hat}''; sentences from \citealp{schoonbaert_2007_RepresentationLexicalSyntactic}).
In crosslingual structural priming, targets that share a grammatical construction with the prime are more likely, even if the two sentences are in different languages (Figure \ref{fig:diagram}).

Human experiments demonstrate robust structural priming effects in a wide variety of languages (\citealp{bock_1986_SyntacticPersistenceLanguage}; see \citet{pickering2008structural} for review) and have been used to argue that bilinguals have shared grammatical representations for their languages.
Structural priming has previously been used to study the structural representations learned by language models \citep{prasad-etal-2019-using,sinclair_2022_StructuralPersistenceLanguage,frank_2021_CrosslanguageStructuralPriming,li-etal-2022-neural,choi2022syntactic2,michaelov2023structural, jumelet-etal-2024-language, zhou2024context}. 
Because the grammatical structure is primed rather than a specific semantic meaning, \citet{sinclair_2022_StructuralPersistenceLanguage} argue that structural priming effects provide evidence for abstract grammatical representations in language models.
By measuring output model probabilities given a prime sentence, structural priming demonstrates causal effects of grammatical representations on model outputs without relying on access to internal model states.
The presence of structural priming in crosslingual scenarios (e.g. a structure primes a similar structure in another language) would indicate representations shared between languages. 

\citet{michaelov2023structural} provided the first evidence for crosslingual structural priming in Transformer language models. The authors argued that this was evidence that language models use shared abstract grammatical representations to represent grammatical constructions for multiple languages.
However, they reported variable and asymmetric effects where for some pairs of languages, structural priming effects were stronger in one direction and weaker (or even non-existent) in the other. 

\paragraph{Language Asymmetries}

In this paper, we first investigate why there are asymmetric effects between languages, i.e. depending on whether they are the target or prime language.  \citet{michaelov2023structural} observe that structural priming effects are stronger when the target language is English. The same effect, observed in human experiments, has been attributed to which language is the first or second learned language (L1 or L2). It is generally thought that structural priming effects are stronger when the prime language is L1 and the target language is L2 (henceforth L1$\rightarrow$L2 priming; \citealp[]{schoonbaert_2007_RepresentationLexicalSyntactic}).
However, a major confound in this line of research is that in most psycholinguistic experiments, English is the L2.
This is due to the populations which are usually sampled from for these experiments, such as university students in countries like the Netherlands \citep[e.g. ][]{schoonbaert_2007_RepresentationLexicalSyntactic,bernolet_2013_LanguagespecificSharedSyntactic}, where it is easiest to find L1 Dutch and L2 English speakers.
If we train language models with different orders of language exposure do we see the same asymmetries? If we do, then the asymmetries are not necessarily due to L1 versus L2, but instead the target language itself (which in this case is English).

\paragraph{Language Similarity}

Second, we investigate whether language similarity impacts structural priming effects. In \citet{michaelov2023structural}, we found more robust structural priming effects for English-Dutch and English-Spanish than for English-Polish and English-Greek sentence pairs. We speculated that this could be in part due to the lower proportions of Polish and Greek training data in the models tested. However, it could also be due to varying language similarity; crosslingual transfer has been shown to be more effective between more similar languages \citep{lin-etal-2019-choosing, ogueji-etal-2021-small, chang-etal-2024-multilinguality}, suggesting a greater degree of representation sharing in similar languages. Polish and Greek are typologically less similar to English than Dutch and Spanish are (\S\ref{sec:lang_similarity}), which might lead to weaker crosslingual structural priming effects.
After controlling for the amount of pre-training data, do differences in the robustness of structural priming correspond to differences in language similarity? If so, then it is possible that asymmetries are not due to order or language exposure, but instead about features of either the prime or target language.

\paragraph{Training Dynamics}
Previous structural priming studies involving language models focus on the final model checkpoint. Here, we ask whether there is a temporal link between the model's acquisition of grammatical knowledge in a second language and its exhibiting structural priming effects. If structural priming effects emerge only after the model learns non-trivial grammatical representations, this reinforces the value of structural priming as a tool for studying multilingual representations.

\paragraph{Contributions}
We train bilingual models, varying the amount of data for each language and the order in which the language model is exposed to each language. With these models, we replicate previous structural priming experiments. We find that asymmetries persist, even when the order of language exposure is reversed. This suggests that asymmetries may be due to features of the target language. Our models also show more robust structural priming effects for more similar language pairs. 
Together, our results not only shed light on shared representations in language models, but may inform our understanding of human structural priming effects.

\section{Related Work}
\label{sec:related-work}

\paragraph{Language Models as Model Organisms}

Our work relates to an ongoing discussion about the role of language models in linguistics and cognitive science \citep{piantadosi2023modern, mahowald2024dissociating, futrell2025linguistics}. 
In a sense, language models are the first \textit{model organism} for language researchers (akin to fruit flies in genetics research), in that they offer the possibility to refine hypotheses about language through the manipulation and evaluation of models, with direct or indirect implications for linguistic theory and related disciplines \citep{muller2024large}. 
For example, in neurolinguistics, \citet{jain2024computational} argue that such \textit{in silico} testing is valuable for evaluating construct validity and refining experiments before they are conducted, as neurolinguistic experiments are extremely costly to run. Similarly, recent work has shown that language models can be valuable model organisms for questions where controlled manipulations are not possible in human experiments. Recent work has used manipulations of training data, for example removing instances of certain grammatical constructions, in order to test questions about language acquisition \citep{patil-etal-2024-filtered, misra-mahowald-2024-language}.
Following this line of reasoning, in this paper, we train language models to have specific L1 and L2 language experience, which would be extremely difficult if not impossible to do with human participants.

\paragraph{Small Models and Syntactic Learning}
Syntax is learned very early in training by language models \citep{blevins-etal-2022-analyzing, chang-etal-2024-characterizing}. Even models trained on human-scaled training data quantities (around 100M words; \citealp[]{warstadt-etal-2023-findings}) show robust syntactic generalizations. Small scale models also permit more manipulations of the training data, given a fixed compute budget, and lend themselves to interpretability analyses.

Training small, controlled models exemplifies the ``controlled rearing" \citep{misra-mahowald-2024-language} approach, in which models are carefully trained with respect to their data exposure in order to make inferences about the effect of training data on model learning. This is only possible for many researchers at a smaller scale.

\paragraph{Bilingual Models}

The bilingual models trained in this paper resemble those in other recent studies using controlled bilingual models to investigate linguistically motivated questions. 
\citet{aoyama-schneider-2024-modeling} first train bilingual models on a first language (L1), freeze some model parameters, then continue training with data from the second language (L2).
\citet{constantinescu2025investigating} train bilingual models with different conditions, similar to our ``interleaved'' and ``simultaneous'' bilingual conditions.

\section{Training Bilingual Language Models} \label{sec:training_models}

We pre-train bilingual language models from scratch to simulate the language experience of bilingual participants in human crosslingual structural priming experiments. We have two conditions. 
In the \textbf{simultaneous bilingual} condition, the models are exposed only to L1 during the first half of training, then an equal mix of L1 and L2 data in the second half. 
Models in the \textbf{sequential bilingual} condition are exposed only to L1 during the first half of training, then only to L2 in the second half. 

We manipulate three factors: language pair (English-Dutch, English-Spanish, English-Polish, English-Greek), language exposure order (e.g. English L1, Dutch L2 vs. Dutch L1, English L2), and bilingual condition (simultaneous or sequential).  This results in a total of 16 language models. For example, we train 4 Dutch models: Dutch-English simultaneous, Dutch-English sequential, English-Dutch simultaneous, and English-Dutch sequential. 
 
Each model is an autoregressive GPT-2 Transformer language model with 124M parameters \citep{radford-etal-2018-improving,radford-etal-2019-language}. 
Following \citet{chang-etal-2024-characterizing}, for each language, we take the first 128M lines of the deduplicated OSCAR corpus \citep{AbadjiOrtizSuarezRomaryetal.2021}.
We train a separate SentencePiece tokenizer \citep{kudo_2018_SentencePieceSimpleLanguage} for each model, using the same language proportions as the model training data.\footnote{For the simultaneous bilingual condition, the overall training data the model sees is 75\% L1 and 25\% L2 data. 
For the sequential bilingual condition, the overall proportions are 50\% L1 and 50\% L2 data.} 
Each sequence is monolingual, but in mixed conditions, batches have alternating L1 and L2 sequences.
We create sequences of 128 tokens, shuffle the sequences, and sample 2B tokens for the training set per language (along with 1M tokens per language for evaluation).
In total, each model is trained for 128,000 steps. Starting at step 64,000, each model is trained on either a mix of L1 and L2 (simultaneous condition) or only L2 data (sequential condition). 
We save checkpoints at regular intervals during training, with increased density just after the introduction of L2, halfway through training. Training details are reported in Appendix \ref{sec:training_details}.
We call these the \textbf{B-GPT models}, and release all checkpoints on Hugging Face.

\subsection{Loss Patterns} \label{sec:model_training}

\begin{figure*}[h!]
\begin{center}
    \includegraphics[width=0.95\linewidth]{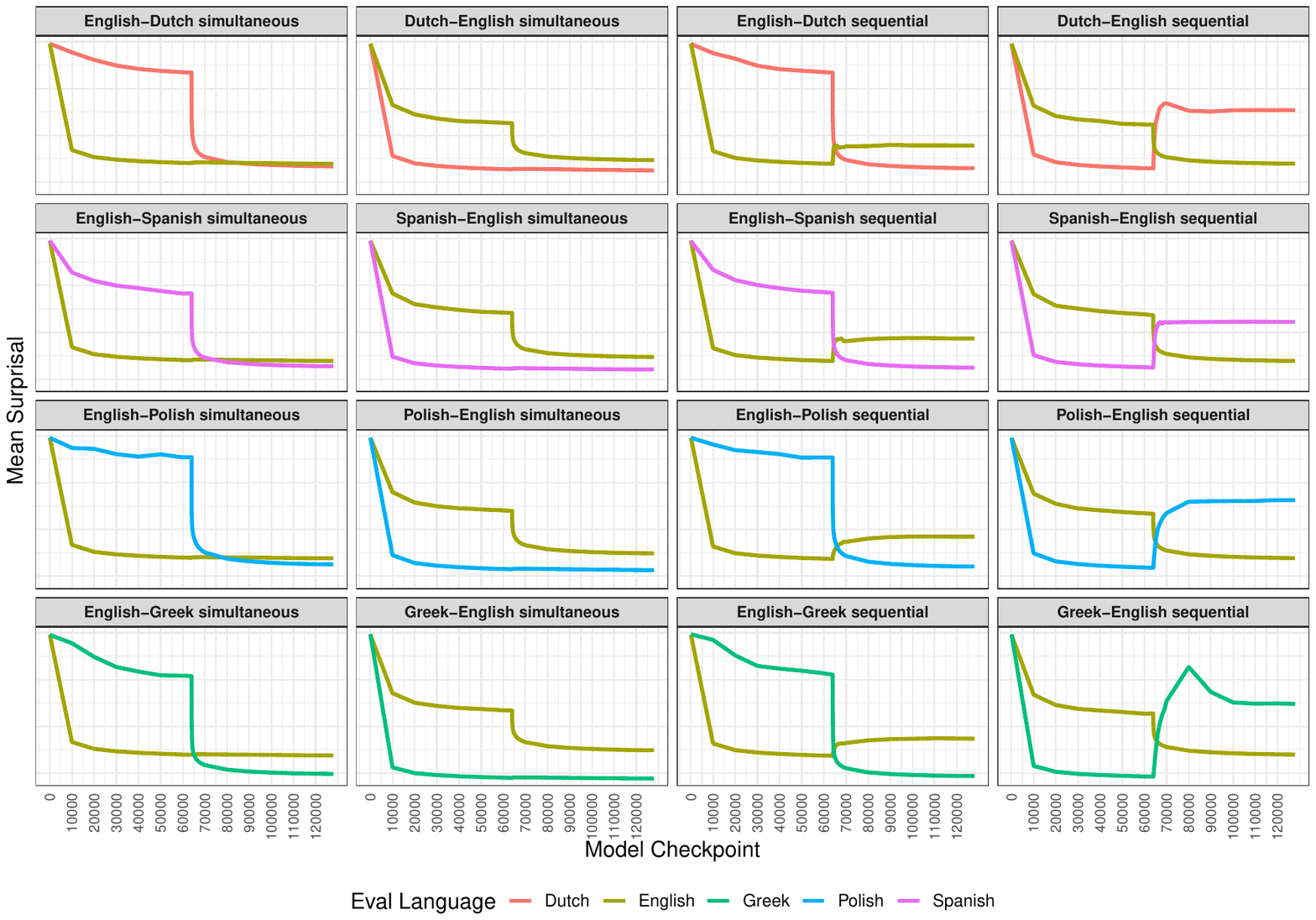}
\end{center}
\caption{L1 and L2 mean surprisal for all models and all checkpoints. The color of each line indicates the evaluation language. Each facet represents one model.}
\label{fig:loss_curves}
\end{figure*}

For each checkpoint, we report the mean surprisal (i.e. log-perplexity or eval loss) on the held out evaluation dataset for both languages each model is trained on (Figure \ref{fig:loss_curves}).
In the simultaneous bilingual condition, we observe consistent patterns: L1 mean surprisal goes down quickly in the first half of training, while L2 mean surprisal stays relatively high. After the introduction of L2 at the halfway point, L2 loss drops quickly. Loss for both languages continues to slowly fall for the rest of training. 
The patterns are quite different for the sequential condition models in the second half of training. After the switch from training on L1 to L2 data, the mean surprisal for the L1 rises sharply.  
Mean surprisal stays high for the rest of training. This is consistent with catastrophic forgetting \citep{mccloskey1989catastrophic}, reflecting the drastic shift in the distribution of training text from L1 to L2.

While all models show similar patterns, the relative mean surprisals do differ somewhat across language pairs.  
For the simultaneous models, especially when English is the L2, there seems to be a language similarity effect. In the second column in Figure \ref{fig:loss_curves}, by the end of training, there is a much smaller difference between mean surprisal for English and Dutch and English and Spanish, relative to the differences in mean surprisal between English and Polish and English and Greek. The lower the mean surprisal for English, the greater the transfer benefit is from the L1. In the case of Dutch, which is the most similar to English of the four languages, the English performance benefits the most. For the Greek-English model, which is typologically and orthographically distinct from English, the English performance gets less of a boost. This is consistent with work showing that linguistic similarity is one of the best predictors for successful crosslingual transfer \citep{chang-etal-2024-multilinguality}.

In the sequential condition, especially when English is the L1 (Figure \ref{fig:loss_curves}, second column from the right), there are differences in the magnitude of the catastrophic forgetting effect. For Dutch the increase in English mean surprisal is less than the increase for the Spanish and Polish, which in turn is less than that for Greek. This also may be due to differences in linguistic similarity (\S\ref{sec:lang_similarity}).

\section{Structural Priming Effects} \label{sec:replication}

We detect structural priming effects by comparing the relative likelihood of a target sentence after different prime sentences, usually pairs of sentences that are semantically identical but vary in their syntax. These are referred to as \textit{grammatical alternations}. 
If a language model assigns a higher probability to a sentence  after a sentence of the same grammatical structure than after a sentence of a different structure, then we consider the language model to exhibit structural priming effects.

\subsection{Calculating Structural Priming Effects}

Following analyses in human studies, structural priming effects are computed as the difference in normalized probability of a target sentence following each prime. 
For example, we compute the normalized probability $P_N$ of a PO target $T_{PO}$ following a PO prime $P_{PO}$ as shown below, where $T_{DO}$ is the DO target and $P_{DO}$ would be a DO prime.

\vspace{-0.25cm}
$$P_N(T_{PO}|P_{PO}) = \frac{P(T_{PO}|P_{PO})}{P(T_{PO}|P_{PO})+P(T_{DO}|P_{PO})}$$

\vspace{0.1cm}

To test for a structural priming effect, we compare $P_N(T_{PO}|P_{PO})$ and $P_N(T_{PO}|P_{DO})$. If the former is significantly higher, i.e. the target following a matching or congruent prime has a higher probability, this would indicate structural priming.
For each model and language combination, we fit a linear mixed effects model predicting the normalized probability of the target with prime type as a fixed effect and experimental item as a random intercept. Here, we only report results for the final model checkpoint, but we conduct the same tests for each model checkpoint. We report the results for the other checkpoints in \S\ref{sec:training_dynamics}. After fitting each linear mixed effects model, we correct for multiple comparisons by controlling for false discovery rate \citep{benjamini_1995_ControllingFalseDiscovery}.

\subsection{Experimental Materials}

We use the experimental stimuli from five studies across the four language pairs, covering three grammatical alternations: DO/PO, s-genitive/of-genitive, and Active/Passive \citep{schoonbaert_2007_RepresentationLexicalSyntactic, bernolet_2013_LanguagespecificSharedSyntactic, hartsuiker_2004_SyntaxSeparateShared, fleischer_2012_SharedInformationStructure, kotzochampou_2022_HowSimilarAre}. We provide further descriptions and examples in Appendix \ref{app:grammatical_alternations}. We check whether the items appear in the training data and report results in Appendix \ref{app:contamination}.

\begin{figure*}[t!]
\begin{center}
\begin{minipage}{\linewidth}
    \includegraphics[width=\linewidth]{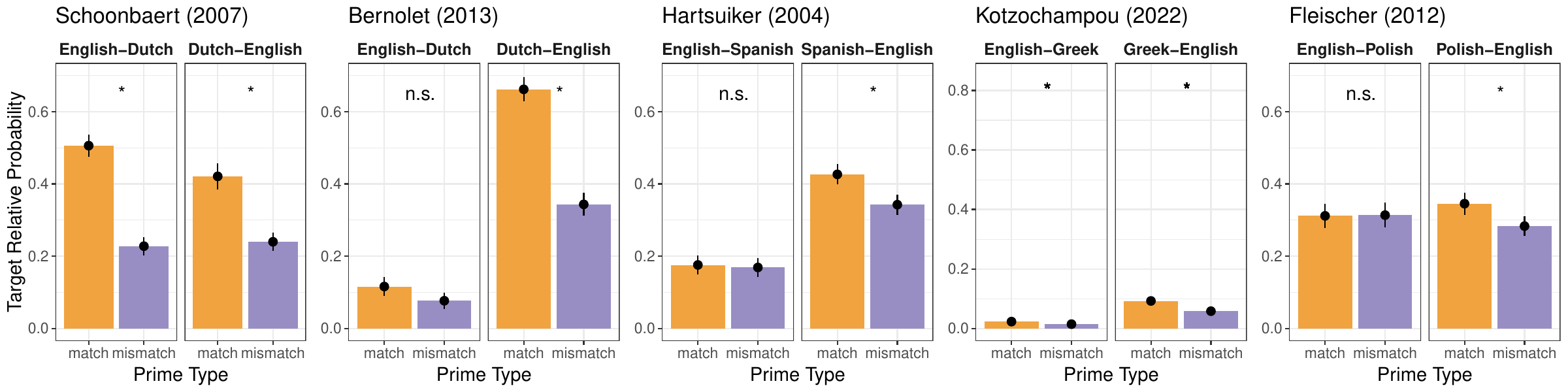}
\end{minipage}

\vspace{0.3cm}  

\begin{minipage}{\linewidth}
    \includegraphics[width=\linewidth]{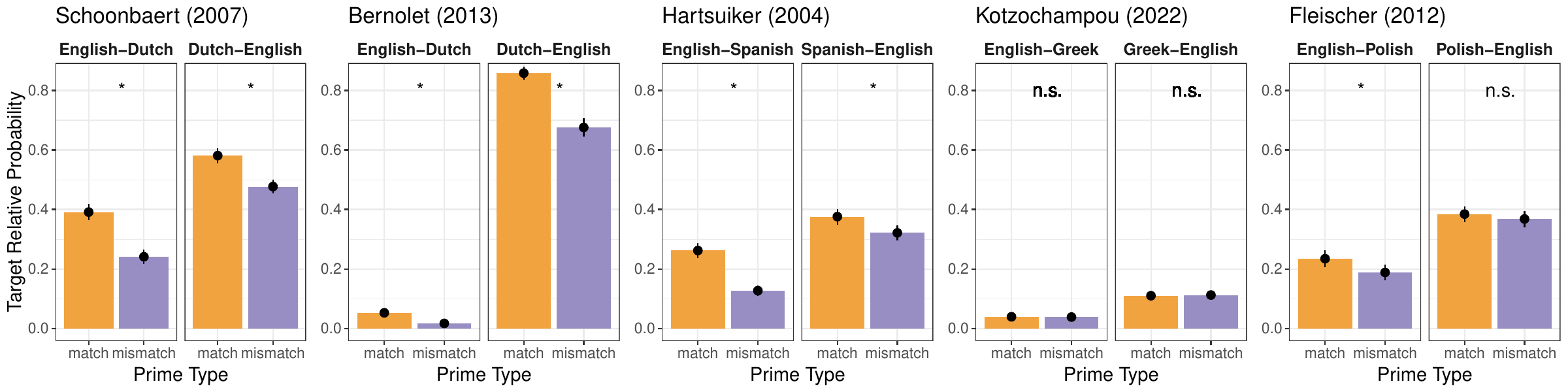}
\end{minipage}
\end{center}

\caption{Priming results for the simultaneous (top) and sequential (bottom) bilingual models. For all experiments, prime language corresponds to L1 and target language corresponds to L2. Significance is indicated with *. Color indicates prime condition. Orange indicates congruent or matching prime and target types and purple indicates mismatched prime and target types. Specific grammatical alternations tested are described in Appendix~\ref{app:grammatical_alternations}.}
\label{fig:replication_combined}
\end{figure*}

\paragraph{DO/PO} The first alternation is for ditransitive events, i.e. events with two objects. One of these is the Prepositional Object (PO) construction (Ex. \ref{po_main}). In this construction, the direct object `hat' directly follows the verb and the indirect object is introduced with a prepositional phrase `to the boxer'. The other is the Double Object (DO) construction (Ex. \ref{do_main}). In this construction, the indirect object `boxer' follows the verb, followed immediately by the direct object `hat'. Dutch has a comparable alternation.

\ex. The cook shows a hat to the boxer. \hfill (PO) \label{po_main}
\vspace{-0.1em}

\ex. The cook shows the boxer a hat. \hfill (DO) \label{do_main} \\
\citep{schoonbaert_2007_RepresentationLexicalSyntactic}

\paragraph{\textit{s}-genitive/\textit{of}-genitive} The second alternation is for genitive constructions, which encode information about possession. In English, one of the constructions is the  s-genitive (Ex. \ref{s_gen_main}), where the possessor `nun' is marked with `'s' and the possessor `nun' precedes the possessed thing `egg'. In the of-genitive construction (Ex. \ref{of_gen_main}), the order is reversed and the preposition `of' is used to express the possessive relationship. Dutch also has this alternation.

\ex. The nun’s egg is yellow. \hfill (s-gen) \label{s_gen_main}
\vspace{-0.1em}

\ex. The egg of the nun is yellow. \hfill (of-gen) \label{of_gen_main} \\
\citep{bernolet_2013_LanguagespecificSharedSyntactic}

\paragraph{Active/Passive} Finally, many events can be encoded using either active or passive voice. In active sentences like Ex. \ref{active_main}, the agent, or do-er of the action, in this case `the taxi', is the syntactic subject of the sentence.  
The theme or patient, i.e. the thing having an action done to it, `truck' in this case, is the syntactic object of the sentence and follows the noun.
In passive sentences, the syntactic subject of the sentence is the theme and the agent is introduced in a prepositional phrase, `by the taxi' (Ex. \ref{passive_main}). 

\ex. The taxi chases the truck. \hfill (Active) \label{active_main}
\vspace{-0.1em}

\ex. The truck is chased by the taxi. \hfill (Passive) \label{passive_main} \\
\citep{hartsuiker_2004_SyntaxSeparateShared}

For each alternation, there are two grammatical constructions which convey the same information and differ primarily in their syntax. For each language pair, both languages share the same grammatical alternation. For example, English and Spanish both share the active/passive alternation. Therefore, for example, we test whether English actives prime Spanish actives and vice versa. 

The original Spanish, Greek, and Polish experiments have many fewer stimuli pairs than the Dutch experiments. Because we do not primarily aim to replicate human experimental results, we create new prime-target pairs by considering every possible pair of prime and target sentences. Then, we randomly sample pairs so that we have 144 pairs each for the Spanish, Greek, and Polish stimuli.
This matches the amount of statistical power for the Dutch experimental materials.

\begin{figure*}[ht!]
    \centering
    \begin{minipage}{0.9\textwidth}
        \centering
        \includegraphics[width=0.9\textwidth]{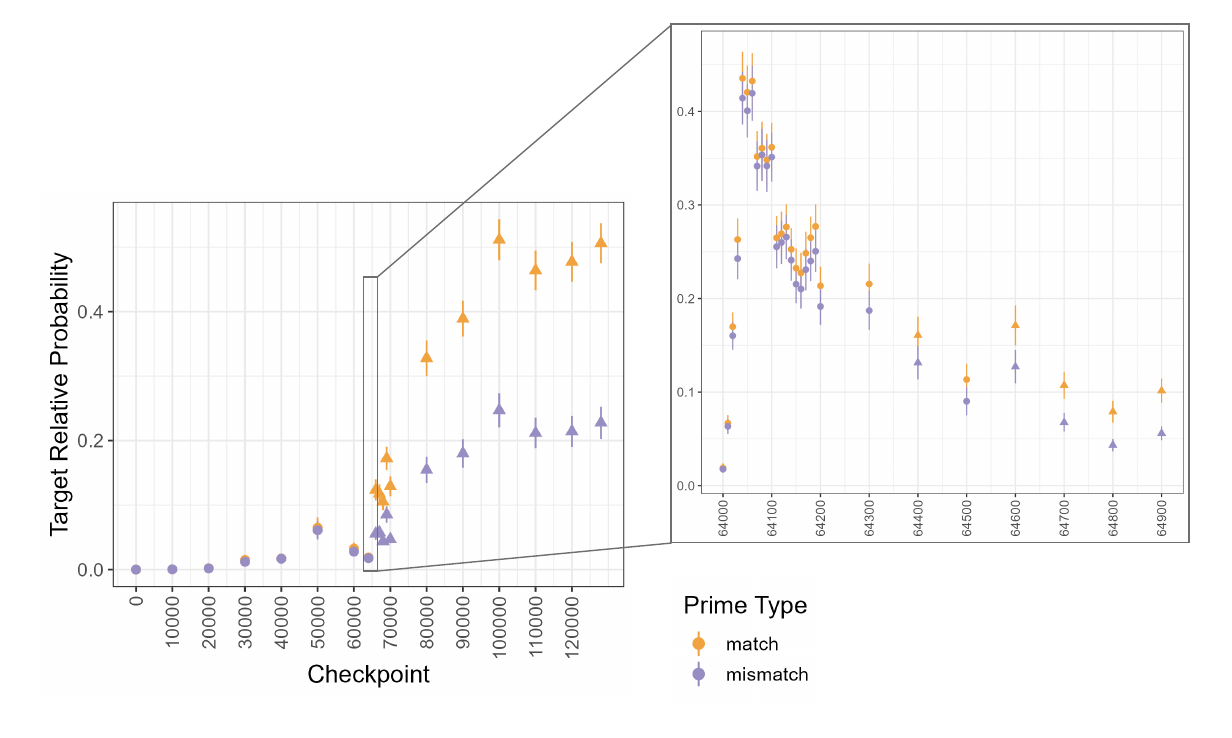}
        \caption{The panel on the left shows structural priming effects for English-Dutch priming for the simultaneous bilingual model, evaluated on \citet{schoonbaert_2007_RepresentationLexicalSyntactic} stimuli. Significant structural priming effects are marked with triangles, and effects that are not significant are marked with circles. In the panel on the right, we plot the structural priming effects for the first 900 steps after L2 exposure. } 
        \label{fig:training_dynamics_example}
    \end{minipage}
\end{figure*}

\subsection{Results} 

Overall, we replicate the crosslinguistic structural priming effects\footnote{Following results from the human structural priming literature, where it has been found that structural priming effects are strongest when the prime language is the participant's L1, and the target language is the L2, we only report results from the L1$\rightarrow$L2 priming conditions. We report L2-L1 priming results in Appendix \ref{app:l2_l1_priming}.} in \citet{michaelov2023structural} (Figure \ref{fig:replication_combined}, top). 
In all cases, when English is the target language, we find that a target sentence is more likely if the prime sentence matches its grammatical structure. 
We also find statistically significant structural priming effects for the experiments with \citet{schoonbaert_2007_RepresentationLexicalSyntactic} and \citet{kotzochampou_2022_HowSimilarAre} stimuli when English is the prime language.
There is still a numerical effect in the expected direction for the experiments with \citet{bernolet_2013_LanguagespecificSharedSyntactic} and \citet{hartsuiker_2004_SyntaxSeparateShared} stimuli where English is the prime language. 

However, there remains an asymmetry in the results, where we see more robust structural priming effects when English is the target language, as opposed to when English is the prime language. We discuss this in depth in Section \ref{sec:asymmetry}.

Notably, we also find structural priming effects in the final checkpoints of the sequential bilingual models (Figure \ref{fig:replication_combined}, bottom) for some languages, despite evidence that the models experienced catastrophic forgetting of L1 (\S\ref{sec:model_training}).
All of the Dutch and Spanish models still exhibit structural priming effects in the final checkpoints, and we see significant structural priming in the English-Polish model. However, there is a reduced effect size, likely caused by the catastrophic forgetting, where L1 knowledge is less well-represented by the end of training despite the fact that shared grammatical representations remain present to some degree.
The stronger effects for Dutch and Spanish, and less strong effects for Greek and Polish, are likely an effect of language similarity with English (\S\ref{sec:lang_similarity}).

\subsection{Training Dynamics} \label{sec:training_dynamics}

\begin{figure}[tb]
    \centering

    \begin{minipage}{0.9\linewidth}
        \centering
        \includegraphics[width=0.95\textwidth]{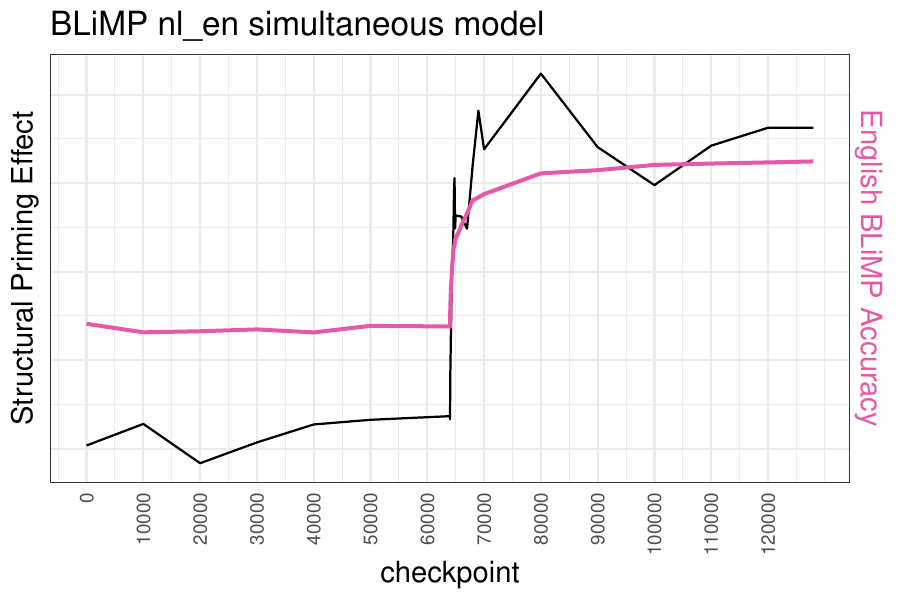}
       
        \vspace{1em}
        
        \includegraphics[width=0.95\textwidth]{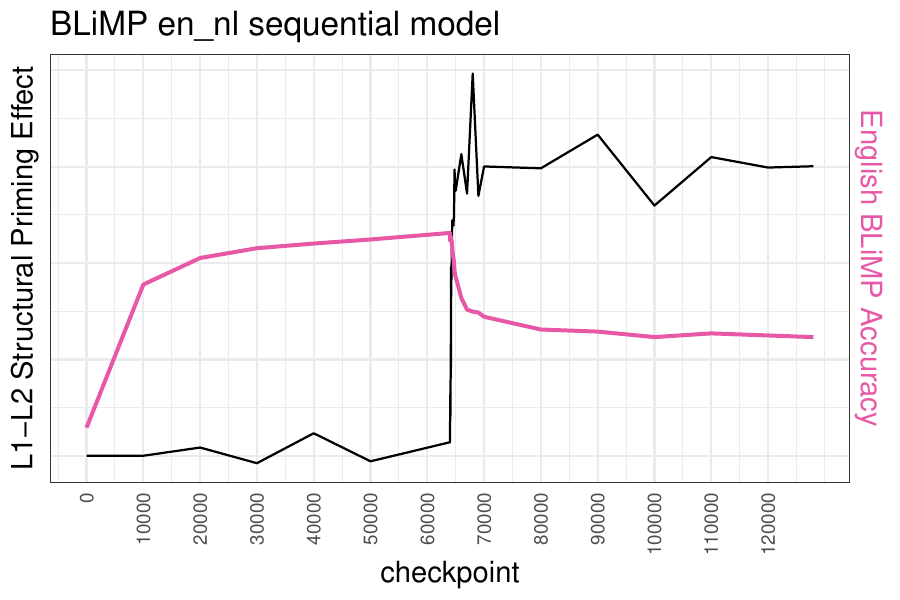} 
    \end{minipage}
    \caption{Structural priming effect and BliMP accuracy over training for Dutch-English simultaneous (top) and English-Dutch sequential (bottom) models.}
    \label{fig:blimp_sp_effect_combined}
\end{figure}

Next, we characterize the time course of the models' learning of shared representations. We first check that structural priming effects are temporally linked to L2 proficiency, because if the models demonstrate structural priming effects before being exposed to L2, we can infer that structural priming is possible through exposure to L1 alone (e.g. due to data contamination across languages). 

To test this, we use BLiMP \citep{warstadt2020blimp} to measure L2 proficiency at each checkpoint.\footnote{While there are BLiMP benchmarks for other languages, BLiMP does not exist for all other languages in our sample. Therefore, we limit our analysis to English BLiMP.} BLiMP measures the grammatical knowledge of the model, which \textcolor{black}{is predictive of a model's ability to generate grammatical text.} We evaluate each model checkpoint on BLiMP using the LM Evaluation Harness \citep{biderman2024lessons}, and we report the average score over all tasks. 
We report results for all models in Appendices \ref{app:blimp_full} and \ref{sec:supplementary_blimp}.

We then evaluate structural priming at each model checkpoint (e.g. Figure \ref{fig:training_dynamics_example} for the English-Dutch simultaneous bilingual model). Before the model is exposed to L2 data, there are no priming effects. But shortly after exposure to L2---as early as 600 steps after exposure to L2, or 4.9M L2 tokens---the language model exhibits stable priming effects. We then compare the time course of structural priming effects to language proficiency. Figure \ref{fig:blimp_sp_effect_combined} 
show structural priming effects as the difference in the relative probabilities between the matching and mismatching prime, plotted in black. In pink, we show the English BLiMP scores.

In the simultaneous bilingual condition (Figure \ref{fig:blimp_sp_effect_combined}, top), structural priming effects emerge at the same time as the model shows a jump in BLiMP performance. Therefore, we argue this draws a stronger link between structural priming behavior and shared multilingual representations. 
In the sequential bilingual condition, we plot L2 English BLiMP accuracy. In the second half of training, accuracy drops as a result of catastrophic forgetting, but structural priming effects still appear and stay relatively high over the course of training. Therefore, it seems that even when the model experiences catastrophic forgetting, representations may still be shared between languages and allow for transfer learning. However, this effect is most clear for Dutch, which is most similar to English. For the other languages, especially Polish and Greek, structural priming effects do not persist after catastrophic forgetting. This is likely another language similarity effect (\S\ref{sec:lang_similarity}). We report comparisons of priming effects and BLiMP accuracy for all models in Appendix \ref{app:blimp_full}.

\section{Discussion}

\subsection{Language Asymmetries} \label{sec:asymmetry}

In human structural priming experiments, it has been shown that structural priming effects are generally stronger in L1$\rightarrow$L2 priming (\citealp[e.g.][]{schoonbaert_2007_RepresentationLexicalSyntactic}), although in some language pairs, there are no priming effects at all. \citet{shin2009syntactic} showed evidence of Korean-English priming, but \citet{shin2011status} found no English-Korean structural priming effects. These experiments have a serious confound, however, as participants are always L2 English speakers. Therefore it is not possible to determine through these experiments whether effect asymmetries are due to L1$\rightarrow$L2 versus L2$\rightarrow$L1 priming or due to the target language being English. In this paper, we found that there were stronger priming effects when English was the target language, independent of its L1/L2 status and when controlling for language exposure. Therefore we argue that the results in the psycholinguistics literature may not be due to differences in L1$\rightarrow$L2 and L2$\rightarrow$L1 priming, but may be driven by whether English is the target language. 

The experiments in this paper rule out the role of model training data quantity, which suggests the asymmetry may be due to cross-linguistic differences.
It is possible that there is something about English as a target language that increases structural priming effects. One candidate is sensitivity to word order. 
In contrast to English, Polish and Greek are morphologically rich languages, where important information is conveyed through morphology (e.g. word inflections), and word orders are less fixed \citep{tzanidaki1995greek,siewierska_1993_SyntacticWeightVs}. Polish and Greek showed less robust structural priming effects across all conditions relative to Dutch and Spanish. Similarly, in human experiments, there is a demonstrated asymmetry for Korean, which also has overt morphological marking and less fixed word order. 
In Tagalog, a language with even more flexible word order, there is evidence from within-language priming that Tagalog speakers do not exhibit structural priming effects based on word order \citep{garcia2020acquisition, garcia2023finding}. 
Therefore, taken together with work in psycholinguistics, the results in this paper call for a reconsideration of the interpretation of previous experimental work. The asymmetries in structural priming effects may be attributed to crosslinguistic differences in the importance of word order, rather than L1/L2 status.

This result serves as an example of the value of language models as model organisms. Disentangling the role of L1$\rightarrow$L2 priming and the role of English as target language is difficult to do with human participants, because it is much easier to find participants for whom English is an L2 than English L1 speakers who speak another language to a high level of proficiency. Our experiments demonstrate the value of language model experiments to develop and refine hypotheses in psycholinguistics that can then be validated through human studies. 

\subsection{Language Similarity} \label{sec:lang_similarity}

In the experiments presented above, there were effects of language similarity throughout. There is a marked difference between the robustness of structural priming effects for Polish and Greek, relative to Dutch and Spanish. In the sequential bilingual condition, the structural priming effects are more robust to catastrophic forgetting when the language pairs are more closely related. In these cases, when we see evidence of catastrophic forgetting, structural priming effects are still present for Dutch and Spanish, but not Greek and Polish. This suggests that in the case of catastrophic forgetting, language similarity is a key factor in the extent to which existing L1 representations will persist after a significant distributional shift in the training data.

As our sample of languages is small and comes from one language family, it is not possible to quantitatively analyze the impact of various typological features. Instead, we explore some possible relevant differences that may affect structural priming effects, such as writing system and how grammatical alternations are encoded.  We provide examples of the alternations and further description in Appendix \ref{app:grammatical_alternations}.
English, Dutch, Spanish, and Polish all use periphrastic constructions to encode the passive voice, whereas Greek uses verbal morphology to do so. 
In English, the difference between the active and passive verb forms is seen in \ref{active_main} and \ref{passive_main}, where the passive is a periphrastic form where the present form of the verb `to be' is combined with the past participle of `chase'. 
By contrast, Greek has a specific verbal morphology to encode active or passive voice. 
This is unlike the other languages included in our experiments, which use a combination of the present copula and the past participle to mark passive voice.

Both of these are typological differences. 
With respect to orthography, Greek is the only language in this set of experiments that uses a non-Latin writing system. Therefore, there is essentially no vocabulary overlap between English and Greek, while the other language pairs may have tokens shared between the languages.
Compounding with typological differences, this differing orthography and lack of shared tokens may contribute to the reduced structural priming effects observed between English and Greek.

By studying shared multilingual representations in language models, our results also tie to work in crosslingual transfer in language models.
\citet{chang-etal-2024-multilinguality} show that language relatedness---especially syntactic typological similarity---is predictive of how much benefit there is to adding multilingual data to improve performance for a target language, relative to a monolingual setting. 
Thus, our results are consistent with previous work showing that crosslingual transfer is more effective between more similar languages. This not only provides a better understanding of crosslingual transfer, but it is indicative of the general limitations of crosslingual transfer. Even for languages in the same language family (in this case, Indo-European), there is still limited ability for models to successfully create shared abstract grammatical representations for language pairs such as Greek and English, relative to a closely related language pair like Dutch and English. 
Therefore, we argue that these results suggest the reconsideration of some current practices for leveraging crosslingual transfer. A common approach for developing a model, especially for a low-resource language, is to start with a powerful open-weight model primarily trained on English and do continued pre-training, vocabulary adaptation, etc. to improve performance for the target language. Our results support previous work showing that using models trained on less data from more similar languages leads to competitive or better results (e.g. \citealp{ogueji-etal-2021-small}).

\section{Conclusion}

In this paper, we used structural priming to understand the shared multilingual representations that drive crosslingual transfer. First, we trained controlled, comparable bilingual language models and replicated crosslingual structural priming effects from previous work. We release the models in order to enable continued work on related questions. We then described the time course of the emergence of structural priming effects relative to the acquisition of L2, drawing a temporal link between L2 proficiency and structural priming effects. We also demonstrated that structural priming effects may persist despite catastrophic forgetting of L1, depending on language similarity between L1 and L2. We argue that language similarity affects several components of this work and should be considered more when attempting to leverage crosslingual transfer in language model development. 

Perhaps most notably, the results in this paper show an asymmetry, where priming effects are stronger when English is the target language. We overcome a confound in prior psycholinguistic research and argue that these results suggest a new interpretation of previous results.

\section*{Limitations}

\paragraph{Language Sample} All of the languages we use in the experiments are Indo-European. While we do cover four distinct sub-branches of the Indo-European language family, this language sample is not sufficiently diverse too draw strong, generalizable conclusions. The language sample is primarily driven by the availability of psycholinguistic datasets, which are more often representative of European languages. 

\paragraph{Model Size} The models we train are very small. This is due to compute limitations. If we trained larger models, we may not have seen the same limits on shared representations and crosslingual transfer, as the models would have not reached capacity limitations as easily. In future follow-up work, increasing the model size would likely be necessary in order to study successful crosslingual transfer in language pairs that are more different than English and Greek or English and Polish. Training larger models and how these effects change with model and data scale would also be illuminating, but is currently not possible given our resources.

Our view is that it is best to first establish evidence for a phenomenon with small models. Now that there is evidence of this phenomenon, we can test larger models in the future to test whether these results change as a function of model scale. And, in this case we also aimed to manipulate several factors. Given our limited compute budget and the fact that we were training models from scratch, we would not have been able to do as many manipulations if we had trained bigger models. Additionally, smaller models more easily allow mechanistic interpretability work, so we feel these models are more useful and accessible at this scale.

\paragraph{Language Data Contamination} While we argue that asymmetries in structural priming effects are due to language differences, it is also possible that the asymmetries could be due to data contamination. If the non-English data could be contaminated with English data, in the cases where English is the target language, the model would see more English data than intended because of contamination. This could boost the structural priming effects, especially when English is the target language. 

Similarly, in Figure \ref{fig:loss_curves}, there is an asymmetry between the English-Dutch and Dutch-English simultaneous models, where the English L2 loss drops much more quickly in the first half of training than does the loss for Dutch as L2. When Dutch is the L1, the model is supposedly not being trained on English. We hypothesize that this is due to English contamination in the Dutch data. The reason we see an asymmetry is likely because there is not as as much Dutch contamination in the English data. This could be due to language use: many Dutch people speak English, but proportionally not as many English speakers also speak Dutch. It could also be due to differences in accuracy of language identification (LID) methods for English and Dutch, as English and Dutch are highly similar languages. 

\section*{Ethical Considerations}

We do not believe the work in this paper raises ethical concerns, but instead we hope it contributes to a better understanding of multilingual language models and indirectly making language models better for more languages. 

We trained 16 small language models. In total, model training took approximately 512 GPU hours on one NVIDIA RTX A6000. The estimated carbon emission for training all models was 66 kg CO$_2$ equivalents.\footnote{Carbon emissions were calculated via \url{https://mlco2.github.io/impact/\#compute}.}
In this paper, we also adhered to the current open science best practices. The training data for our language models is available and falls under fair use. The code to train and evaluate the models is available.\footnote{\url{https://osf.io/5cw2e/}}
The experimental stimuli from \citet{schoonbaert_2007_RepresentationLexicalSyntactic}, \citet{bernolet_2013_LanguagespecificSharedSyntactic}, \citet{hartsuiker_2004_SyntaxSeparateShared}, \citet{fleischer_2012_SharedInformationStructure}, and \citet{kotzochampou_2022_HowSimilarAre} are scientific research materials, and as such, we believe that their use for scientific research falls under the category of fair use.
We release the language models we trained under an Apache 2.0 license, which allows for modification and distribution with minimal restrictions.

\section*{Acknowledgements}
We would like to thank 
Sarah Bernolet, 
Vasiliki Chondrogianni, 
Zuzanna Fleischer, 
Robert J. Hartsuiker, 
Sotiria Kotzochampou, 
Janet F. McLean, 
Martin J. Pickering, 
Sofie Schoonbaert, 
and Eline Veltkamp 
for making their
experimental stimuli available; and Nikitas Angeletos Chrysaitis, Pamela D. Rivière Ruiz, 
Quirine van Engen, Alexandra Taylor,
Robert Slawinski, 
Tiffany Wu, Fiona Tang, Emily Xu, and Jason Tran
for their assistance in preparing them for use in the
present study.
Models were pre-trained and evaluated using hardware provided by the NVIDIA Corporation as part of an NVIDIA Academic Hardware Grant.
Tyler Chang is partially supported by the UCSD HDSI graduate fellowship.
James Michaelov was supported by a grant from the Andrew W. Mellon foundation (\#2210-13947) during the writing of this paper.

\bibliography{custom}

\appendix

\section{Grammatical Alternations} \label{app:grammatical_alternations}
\renewcommand{\thetable}{A.\arabic{table}}
\renewcommand{\thefigure}{A.\arabic{figure}}

\paragraph{DO/PO} We use the Dutch and English stimuli from \citet{schoonbaert_2007_RepresentationLexicalSyntactic}, which contain pairs that contrast the Prepositional Object (PO) and Double Object (DO) dative constructions. 

In some languages, for ditransitive sentences, when there are two objects, there are two possible ways to express the same event. One of these is the \textbf{Prepositional Object (PO)} construction (Ex. \ref{po}). In this construction, the direct object `hat' directly follows the verb and the indirect object is introduced with a prepositional phrase `to the boxer'. The other is the \textbf{Double Object (DO)} construction (Ex. \ref{do}). In this construction, the indirect object `boxer' follows the verb, followed immediately by the direct object `hat'.

\ex. 
\a. The cook shows a hat to the boxer. \hfill (PO) \label{po}
\b. The cook shows the boxer a hat. \hfill (DO) \label{do}
\b.[] \citep{schoonbaert_2007_RepresentationLexicalSyntactic}

Dutch has an equivalent alternation, with the same word order as English for PO (Ex. \ref{nl_po}) and DO (Ex. \ref{nl_do}) sentences.

\ex.
\ag. De kok toont een hoed aan de bokser.  \label{nl_po}\\ 
The cook shows a hat to the boxer.  \\
\bg. De kok toont de bokser een hoed.   \\
The cook shows the boxer a hat. \label{nl_do} \\ %
\b.[] \citep{schoonbaert_2007_RepresentationLexicalSyntactic}

\paragraph{\textit{s}-genitive/\textit{of}-genitive}
We use the Dutch and English stimuli from \citet{bernolet_2013_LanguagespecificSharedSyntactic}, which contrast the two genitive constructions, which are semantically equivalent ways to express possession. In English, one of these is the  \textbf{s-genitive} construction (Ex. \ref{s_gen}), where the possessor `nun' is marked with `'s'. In this construction, the possessor `nun' precedes the possessed thing `egg'. In the \textbf{of-genitive} construction (Ex. \ref{of_gen}), the order is reversed and the possessed thing precedes the possessor. In this case, the preposition `of' is used to express the possessive relationship. 

\ex.
\a. The nun’s egg is yellow. \hfill (s-gen) \label{s_gen}
\b. The egg of the nun is yellow. \hfill (of-gen) \label{of_gen}
\b.[] \citep{bernolet_2013_LanguagespecificSharedSyntactic}

Dutch has a similar alternation. For proper names, s-genitive possession can be marked with `'s', but for common nouns, possession is marked with the possessive pronoun that corresponds in gender to the possessor noun. In the example below (Ex. \ref{nl_s_gen}), \textit{non} `nun' is feminine, so \textit{haar} `her' marks possession. Masculine nouns use \textit{zijn} `his' \citep{bernolet_2013_LanguagespecificSharedSyntactic}. The Dutch of-genitive construction is more similar to English, where the preposition \textit{van} `of' is used to show possession, and the order of the possessor and possessee is flipped, relative to the s-genitive order.

\ex. 
\ag. De non haar ei is geel. \label{nl_s_gen}\\
The nun \textsc{POSS} egg is yellow. \\
\bg. Het ei van de non is geel. \label{nl_of_gen} \\
The egg of the nun is yellow. \\
\b.[] \citep{bernolet_2013_LanguagespecificSharedSyntactic}

\paragraph{Active/Passive} For Spanish-English, Polish-English, and Greek-English experiments, we use stimuli that contrast active and passive constructions. For Spanish-English, we use stimuli from \citet{hartsuiker_2004_SyntaxSeparateShared}; for Greek-English, the stimuli come from \citet{kotzochampou_2022_HowSimilarAre}; and for Polish-English, we use stimuli from \citet{fleischer_2012_SharedInformationStructure}.

Many languages allow events to be expressed as either active or passive. In \textbf{active} sentences, e.g. Ex. \ref{active}, the agent, or do-er of the action, `the taxi' is the syntactic subject of the sentence, which in English, is  marked by being the first argument in the sentence.  
The theme or patient, i.e. the thing having an action done to it, `truck' is the syntactic object of the sentence and follows the noun.
In \textbf{passive} sentences, the syntactic subject of the sentence is the theme. The agent is introduced in a prepositional phrase, `by the taxi' (Ex. \ref{passive}). 

\ex. 
\a. The taxi chases the truck. \hfill (Active) \label{active}
\b. The truck is chased by the taxi. \hfill (Passive) \label{passive}
\b.[] \citep{hartsuiker_2004_SyntaxSeparateShared}

Spanish expresses active and passive sentences very similarly to English, following the same word order (Ex. \ref{es_active} and \ref{es_passive}, respectively).

\ex. 
\ag. El taxi persigue el camión. \label{es_active}\\
The taxi chases the truck. \\
\bg. El camión es perseguido por el taxi. \label{es_passive}\\
The truck is chased by the taxi.\\
\b.[] \citep{hartsuiker_2004_SyntaxSeparateShared}

Typologically, Polish and Greek are more different from English than either Dutch or Spanish is. Both of these languages mark the syntactic subjects and objects using case marking, unlike English, Dutch, and Spanish, which do this only with word order. In Polish, for example, in the active, \textit{sportowiec} `sportsman' is in the nominative case and is the syntactic subject of the sentence.  The patient `ballet dancer' takes the accusative and is the grammatical object of the sentence. In the passive, it is in the accusative case (\textit{sportowca}) and is introduced with a prepositional phrase. The patient  `ballet dancer', in this case, is in the nominative case.

\ex. 
\ag. Sportowiec przygniata baletnic\k{e}.  \label{pol_active_1}  \\
sportsman.NOM.SG squash.PRES.3SG ballet-dancer.ACC.SG   \\
"The sportsman squashes the ballet dancer." 
\bg. Baletnica jest przygniatana przez sportowca.\label{pol_passive_1}\\
ballet-dancer.NOM.SG be.3SG.PRES squash.PST.PART by sportsman.ACC.SG\\
"The ballet dancer is squashed by the sportsman." 
\b.[] \citep{fleischer_2012_SharedInformationStructure}

Similarly, Greek marks subject and object roles with case marking. When it is the subject, \textgreek{αθλητής} (\textit{athlitis}) `athlete' is nominative, but as an object, it takes the accusative case (\textgreek{αθλητή}, \textit{athliti}). 
Greek, unlike Polish or the other languages described here, has a specific verbal morphology to encode active or passive voice (compare \ref{el_active} and \ref{el_passive}), therefore the verb form is also specific to passive voice, unlike the other languages shown here, which use a combination of the present copula and the past participle to mark passive voice.

\ex.
\ag. \textgreek{Ο} \textgreek{αθλητής} \textgreek{κλωτσ\textbf{άει}} \textgreek{τον} \textgreek{κλέφτη.} \label{el_active} \\
O athlitis klots\textbf{aei} ton klefti. \\
The athlete.NOM kicks-\textsc{ACTIVE} the thief.ACC. 
\b.[] "The athlete kicks the thief." 
\bg. \textgreek{Ο} \textgreek{κλέφτης} \textgreek{κλωτσ\textbf{ιέται}} \textgreek{από} \textgreek{τον} \textgreek{αθλητή}. \label{el_passive}\\
O kleftis klots\textbf{iete} apo ton athliti. \\
The thief.NOM kicks-\textsc{\textbf{PASSIVE}} by the athlete.ACC. 
\b.[] "The thief is kicked by the athlete." 
\b.[] \citep{kotzochampou_2022_HowSimilarAre}

\section{Contamination Analysis} \label{app:contamination}
\renewcommand{\thetable}{B.\arabic{table}}
\renewcommand{\thefigure}{B.\arabic{figure}}

The ``What's in My Big Data?'' tool\footnote{\url{https://wimbd.apps.allenai.org/}} indexes OSCAR and allows $n$-gram search, but we were unable to access it. Instead, we use Infini-gram\footnote{\url{https://huggingface.co/spaces/liujch1998/infini-gram}} \citep{liuinfini}, which indexes C4, which is also compiled of multilingual web data and is much larger than the portion of OSCAR we used to train our models. 
Only a very small number of our stimuli can be found in C4 (Table~\ref{tab:contamination}).

\begin{table}[h!]
\centering
\begin{tabular}{@{}lll@{}}
\toprule
              & \multicolumn{2}{l}{\textbf{Contaminated Items}} \\ \midrule
\textbf{Dataset}       & \textbf{Count}           & \textbf{Proportion }          \\ \midrule
Schoonbaert   & 0               & 0                    \\
Bernolet      & 0               & 0                    \\
Hartsuiker    & 3               & 0.0078               \\
Fleischer     & 0               & 0                    \\
Kotzochamopou & 4               & 0.0208               \\ \bottomrule
\end{tabular}
\caption{C4 contamination results.}
\label{tab:contamination}
\end{table}

\section{Model Training Details} \label{sec:training_details}
\renewcommand{\thetable}{C.\arabic{table}}
\renewcommand{\thefigure}{C.\arabic{figure}}

Model training code is based on that from \citet{chang2022word}.\footnote{Available at \url{https://github.com/tylerachang/word-acquisition-language-models}.}

\paragraph{Model Hyperparameters}

Table \ref{tab:hyperparams} shows the model training hyperparameters.

\setlength\tabcolsep{3pt}
\begin{table}[hb!]
    \centering
    \renewcommand{\arraystretch}{1.11}
    \begin{tabular}{|p{5cm}|r|}
        \hline
        \textbf{Hyperparameter} & \textbf{Value} \\
        \hline
        Layers & 12 \\
        Embedding size & 768 \\
        Hidden size & 768 \\
        Intermediate hidden size & 3072 \\
        Attention heads & 12 \\
        Attention head size & 64 \\
        Activation function & GELU \\
        Vocab size & 50004 \\
        Max sequence length & 128 \\
        Position embedding & Absolute \\
        Batch size & 128 \\
        Train steps & 128k \\
        Learning rate decay & Linear \\
        Warmup steps & 10000 \\
        Learning rate & 1e-4 \\
        Adam $\epsilon$ & 1e-6 \\
        Adam $\beta_1$ & 0.9 \\
        Adam $\beta_2$ & 0.999 \\
        Dropout & 0.1 \\
        Attention dropout & 0.1 \\
        \hline
    \end{tabular}
    \normalsize
    \caption{Language model hyperparameters.}
    \label{tab:hyperparams}
\end{table}

\paragraph{Checkpoints}

We take checkpoints at the first and last steps (128k). Additionally we take checkpoints every 10k steps. After the introduction of the L2 at the halfway point (64k), we save checkpoints every 10 steps, because we expect that structural priming effects may emerge within the first few hundred training steps after the introduction of L2. 
After 200 steps after the introduction of L2, we gradually increase the checkpoint intervals. This way, we have increased resolution during the period of training where we expect to see the emergence of structural priming effects, while minimizing the number of checkpoints needed. 

We save model checkpoints at the following training steps: 0, 10000, 20000, 30000, 40000, 50000, 60000, 64000, 64010, 64020, 64030, 64040, 64050, 64060, 64070, 64080, 64090, 64100, 64110, 64120, 64130, 64140, 64150, 64160, 64170, 64180, 64190, 64200, 64300, 64400, 64500, 64600, 64700, 64800, 64900, 65000, 66000, 67000, 68000, 69000, 70000, 80000, 90000, 100000, 110000, 120000, 128000.

\section{L2-L1 Priming} \label{app:l2_l1_priming}
\renewcommand{\thetable}{D.\arabic{table}}
\renewcommand{\thefigure}{D.\arabic{figure}}

\begin{figure*}[h!]
\begin{center}
    \includegraphics[width=\linewidth]{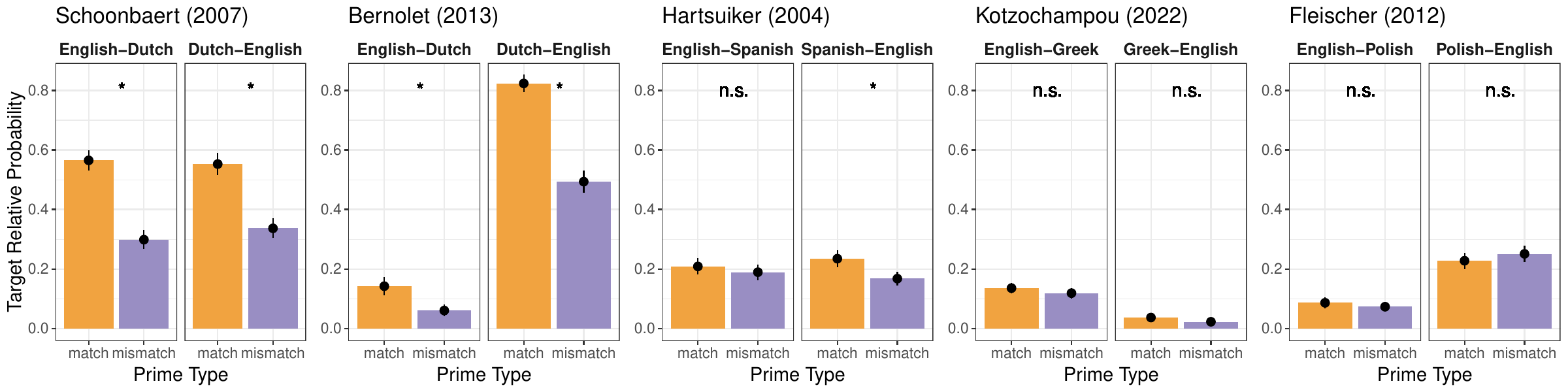}
\end{center}
\caption{Simultaneous bilingual condition. Prime language corresponds to L2.}
\label{fig:app_l2l1_simulaneous}
\end{figure*}

\begin{figure*}[h!]
\begin{center}
    \includegraphics[width=\linewidth]{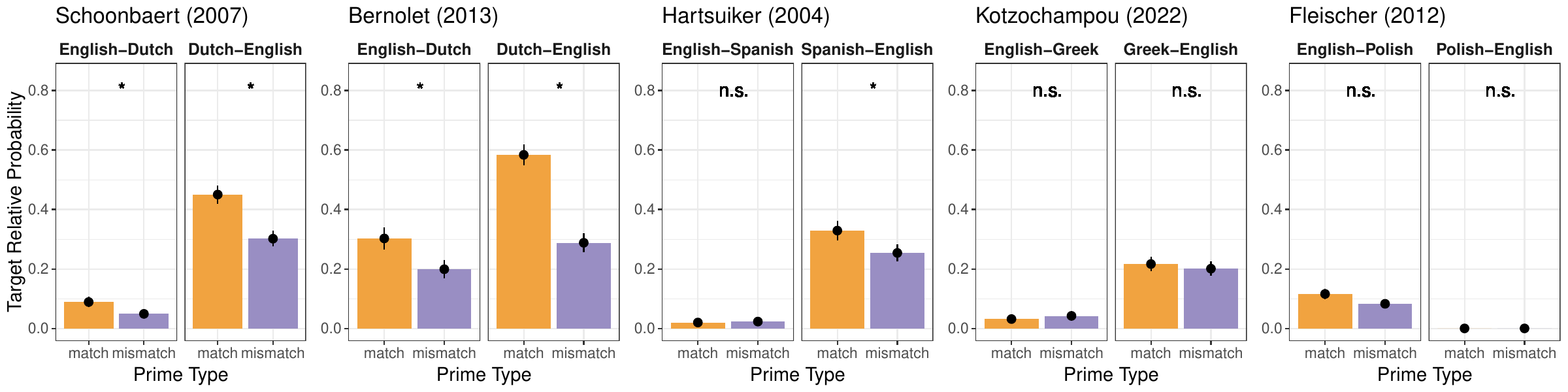}
\end{center}
\caption{Sequential bilingual condition. Prime language corresponds to L2.}
\label{fig:app_l2l1_sequential}
\end{figure*}

Figures \ref{fig:app_l2l1_simulaneous} and \ref{fig:app_l2l1_sequential} show the L2$\rightarrow$L1 results for all models for both the simultaneous and sequential bilingual conditions, respectively. Each facet represents a model. The labels, e.g. English-Dutch and Dutch-English, correspond to the L1 and L2 of each model. 

\section{Full BLiMP Results} \label{app:blimp_full}
\renewcommand{\thetable}{E.\arabic{table}}
\renewcommand{\thefigure}{E.\arabic{figure}}

\subsection{Schoonbaert (2007)}

\begin{figure*}[h]
    \centering
    \begin{minipage}[b]{0.45\linewidth}
        \centering
        \includegraphics[width=\linewidth]{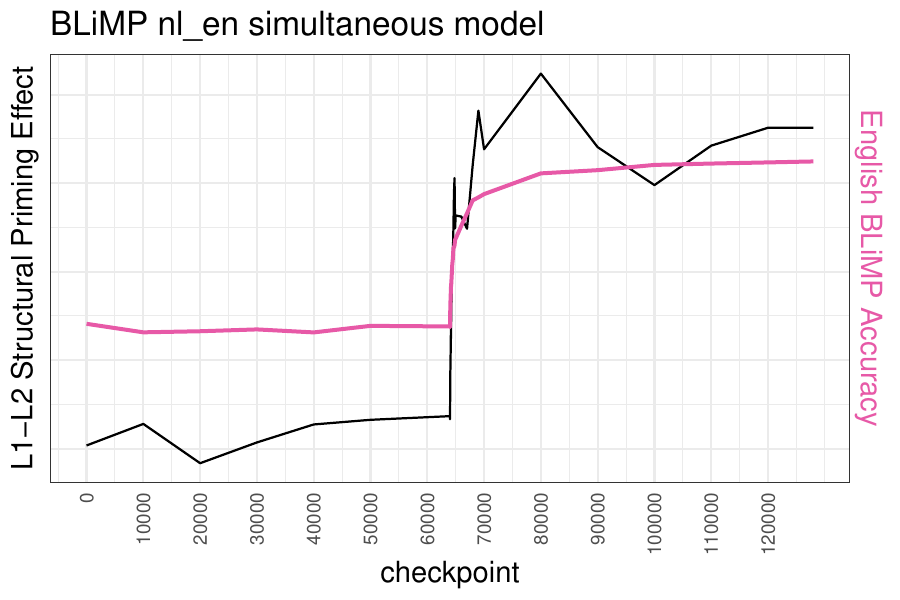} 
        \vspace{0.5em}
        \includegraphics[width=\linewidth]{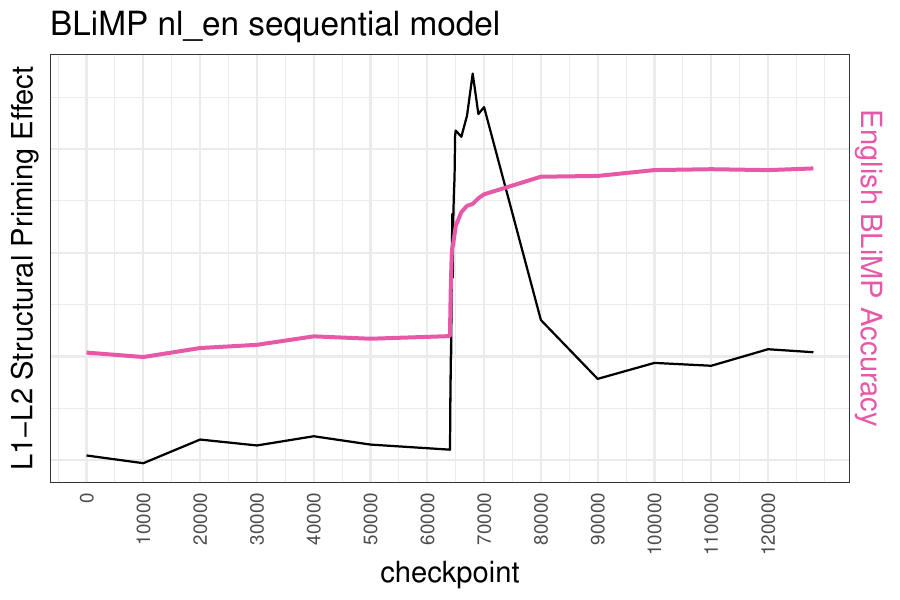} 
        \vspace{0.5em}
    \end{minipage}
    \hfill 
    \begin{minipage}[b]{0.45\linewidth}
        \centering
        \includegraphics[width=\linewidth]{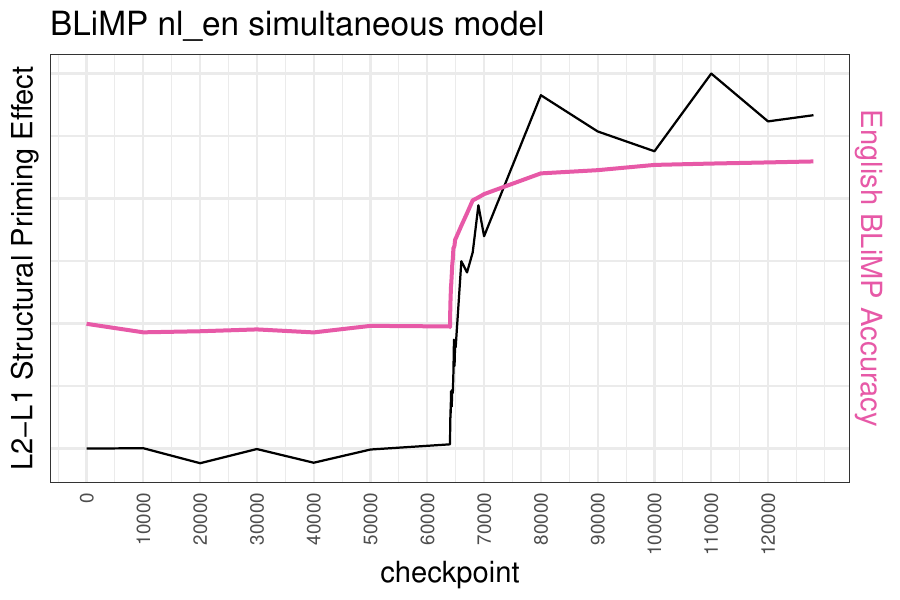} 
        \vspace{0.5em}
        \includegraphics[width=\linewidth]{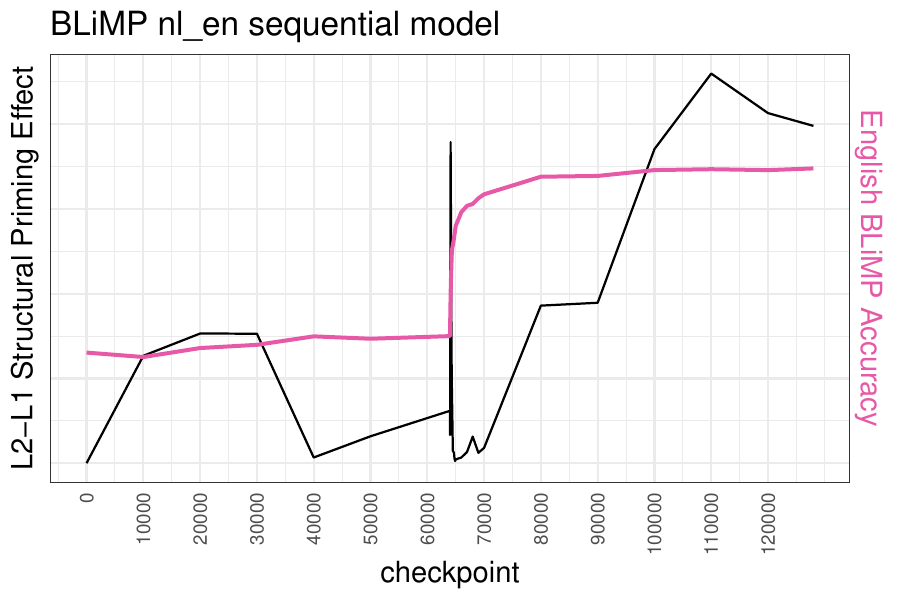} 
        \vspace{0.5em}
    \end{minipage}
    \caption{Structural priming effect (black), plotted as the difference between match and mismatch conditions, and English BLiMP accuracy (pink) over the course of model training. Y-axes have been re-scaled for easier comparison.}
    \label{fig:schoonbaert_blimp_full}
\end{figure*}

Figure \ref{fig:schoonbaert_blimp_full} shows the comparison for structural priming effects and BLiMP scores for all Dutch-English models. 

\subsection{Bernolet (2013)}

\begin{figure*}[h]
    \centering
    \begin{minipage}[b]{0.45\linewidth}
        \centering
        \includegraphics[width=\linewidth]{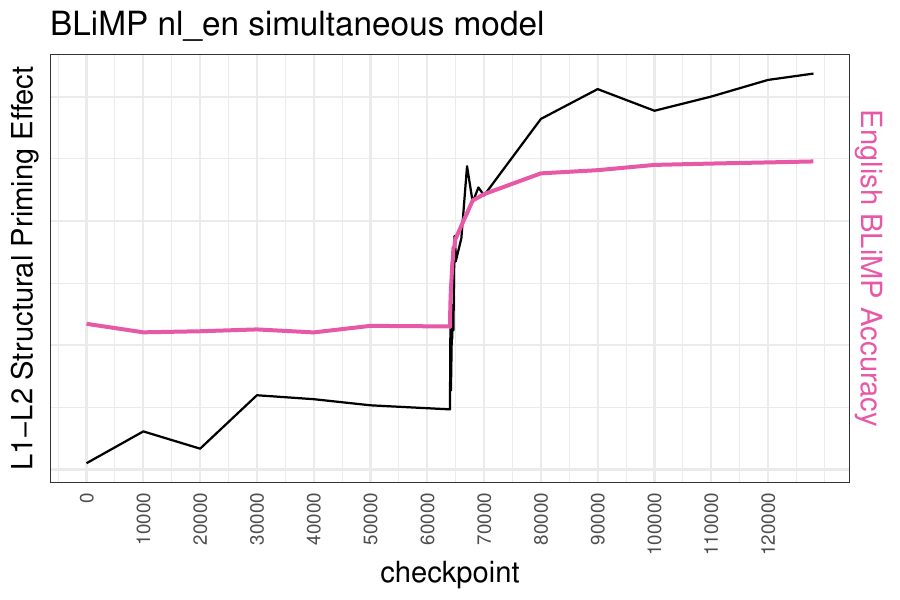}
        \vspace{0.5em}
        \includegraphics[width=\linewidth]{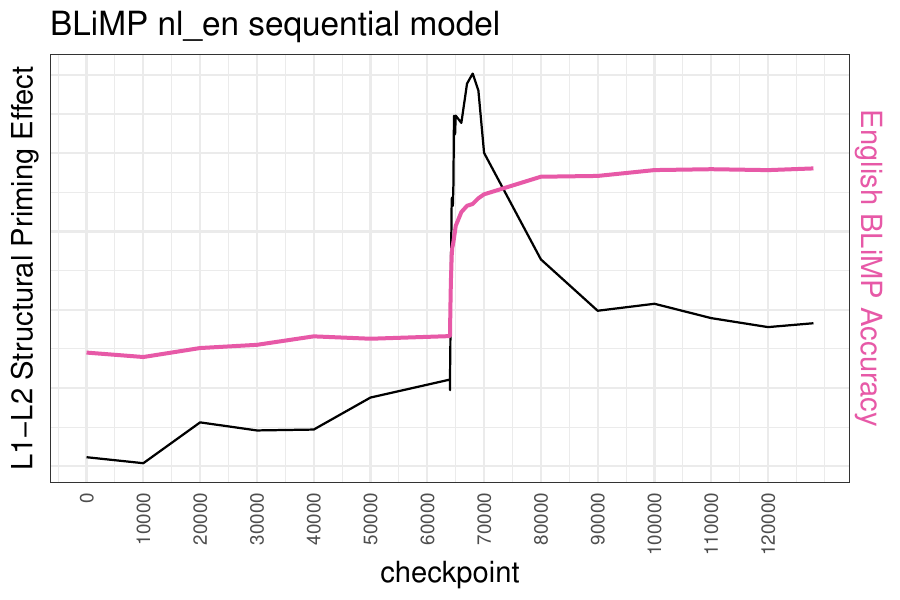} 
        \vspace{0.5em}
    \end{minipage}
    \hfill
    \begin{minipage}[b]{0.45\linewidth}
        \centering
        \includegraphics[width=\linewidth]{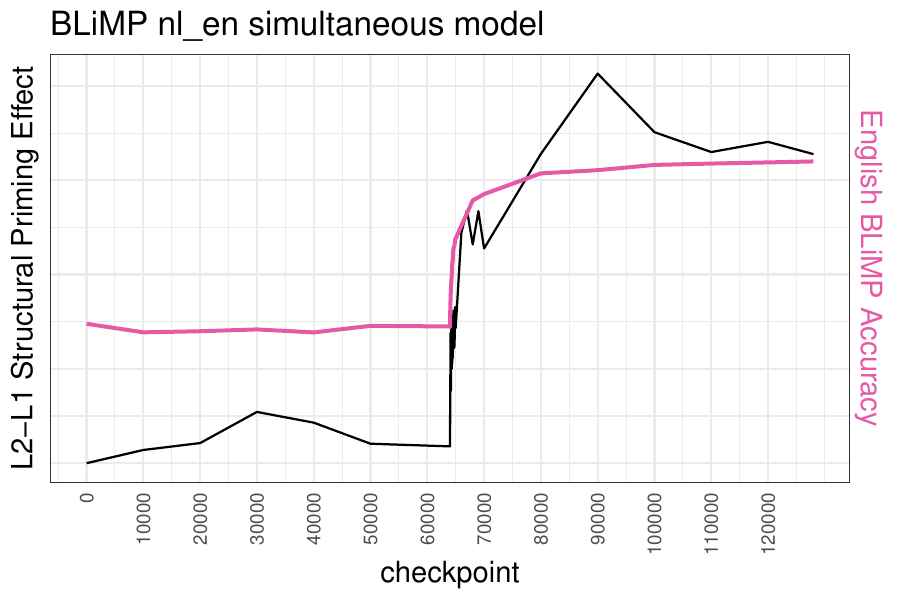} 
        \vspace{0.5em}
        \includegraphics[width=\linewidth]{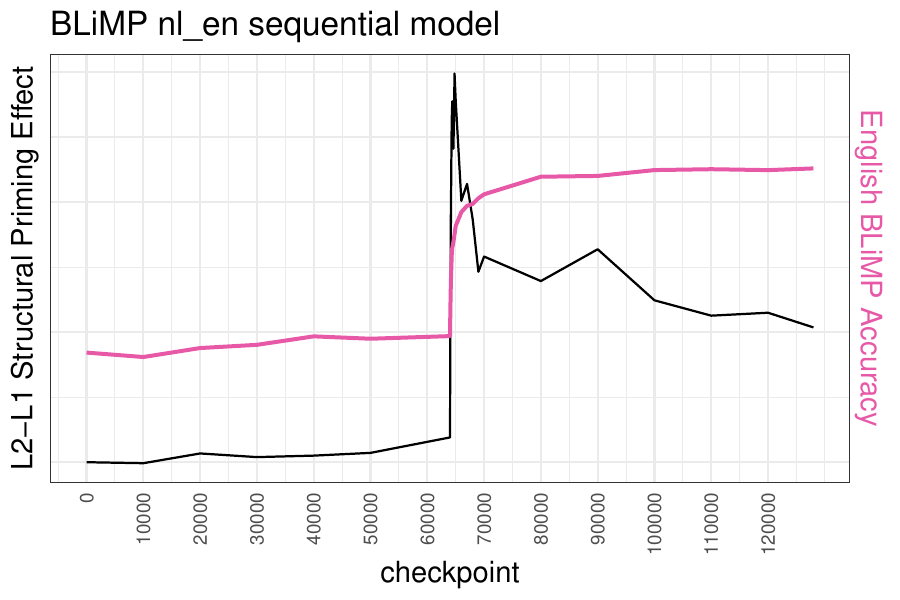} 
        \vspace{0.5em}
    \end{minipage}
    \caption{Structural priming effect (black), plotted as the difference between match and mismatch conditions, and English BLiMP accuracy (pink) over the course of model training. Y-axes have been re-scaled for easier comparison.}
    \label{fig:bernolet_blimp_full}
\end{figure*}

Figure \ref{fig:bernolet_blimp_full} shows the comparison for structural priming effects and BLiMP scores for all Dutch-English models. 

\subsection{Hartsuiker (2004)}

\begin{figure*}[h]
    \centering
    \begin{minipage}[b]{0.45\linewidth}
        \centering
        \includegraphics[width=\linewidth]{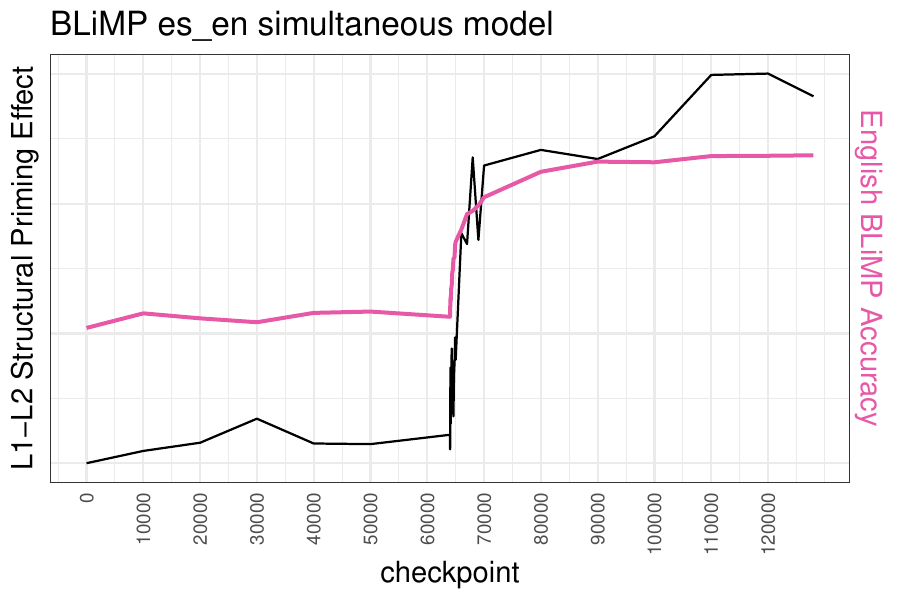} 
        \vspace{0.5em}
        \includegraphics[width=\linewidth]{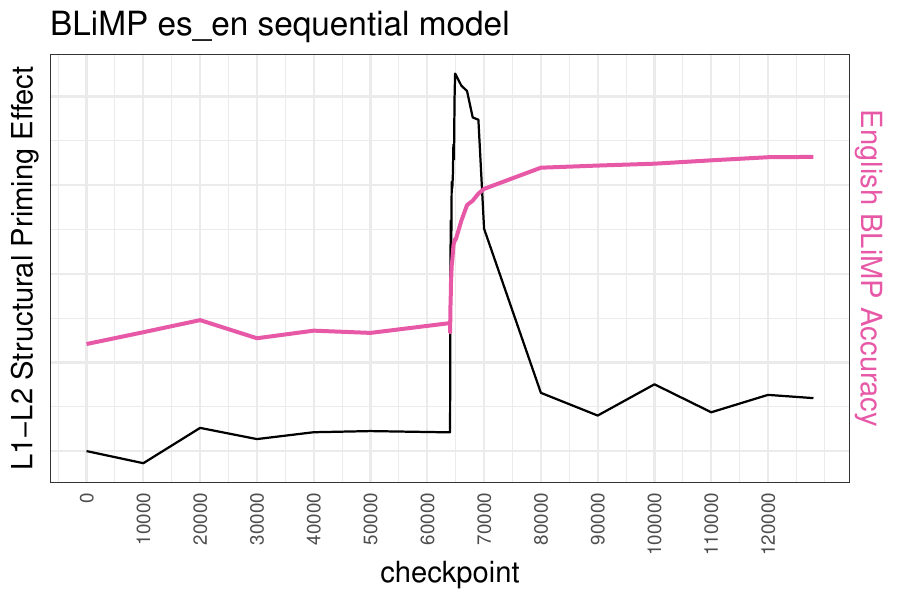} 
        \vspace{0.5em}
    \end{minipage}
    \hfill 
    \begin{minipage}[b]{0.45\linewidth}
        \centering
        \includegraphics[width=\linewidth]{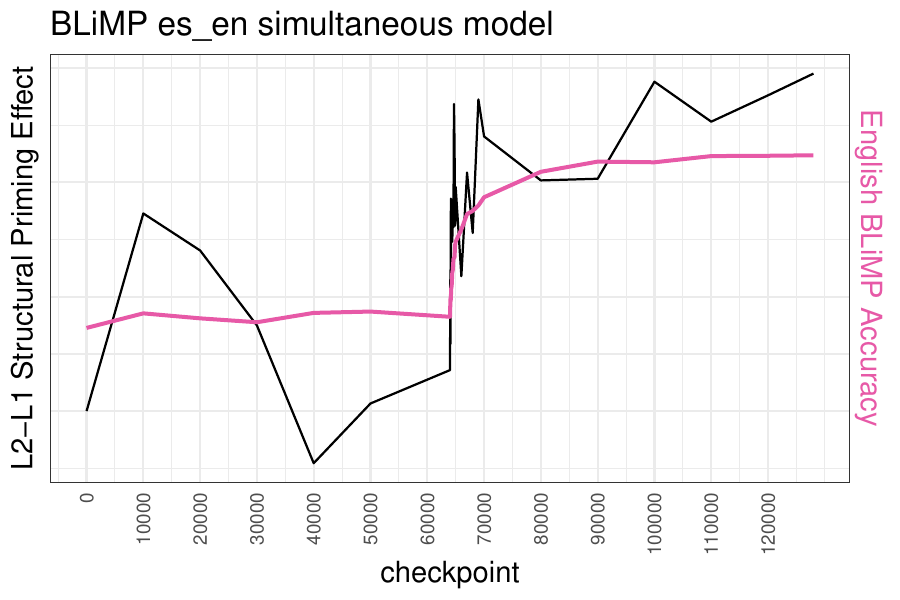} 
        \vspace{0.5em}
        \includegraphics[width=\linewidth]{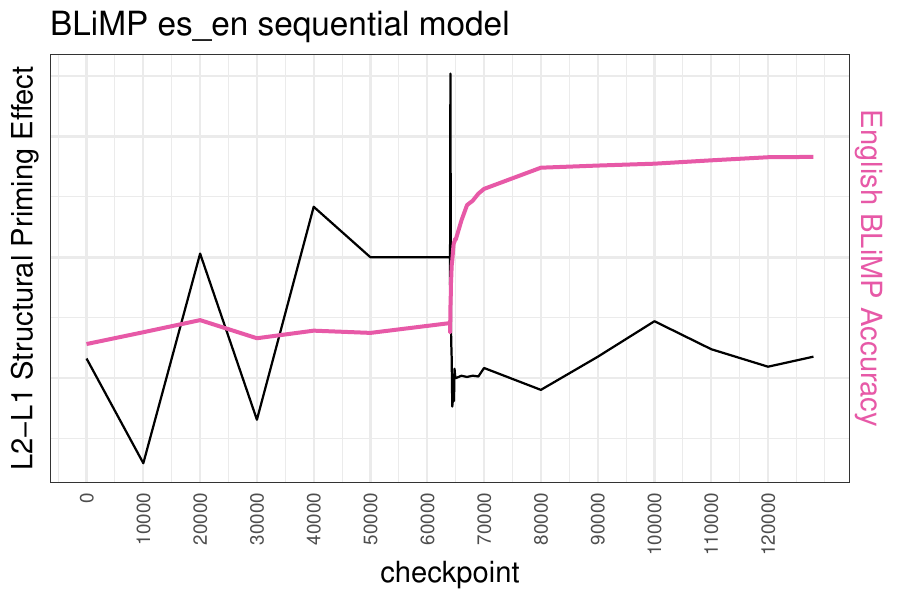} 
        \vspace{0.5em}
    \end{minipage}
    \caption{Structural priming effect (black), plotted as the difference between match and mismatch conditions, and English BLiMP accuracy (pink) over the course of model training. Y-axes have been re-scaled for easier comparison.}
    \label{fig:hartsuiker_blimp_full}
\end{figure*}

Figure \ref{fig:hartsuiker_blimp_full} shows the comparison for structural priming effects and BLiMP scores for all Spanish-English models. 

\subsection{Fleischer (2012)}

\begin{figure*}[h]
    \centering
    \begin{minipage}[b]{0.45\linewidth}
        \centering
        \includegraphics[width=\linewidth]{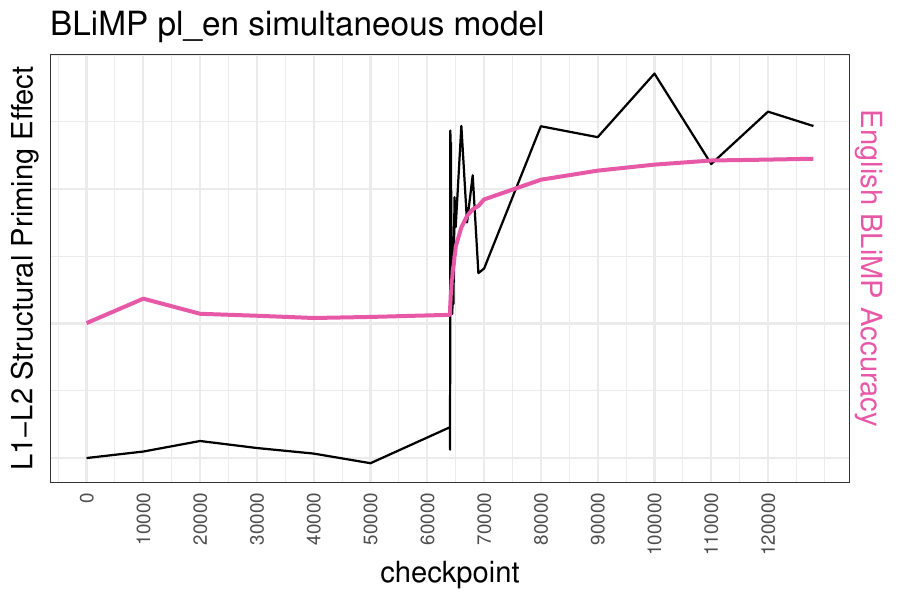} 
        \vspace{0.5em}
        \includegraphics[width=\linewidth]{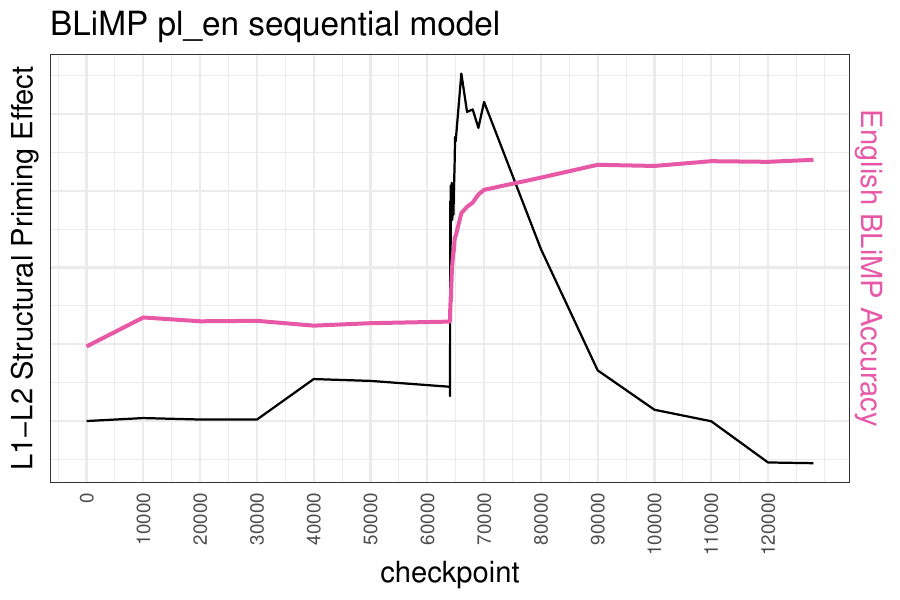}
        \vspace{0.5em}
    \end{minipage}
    \hfill 
    \begin{minipage}[b]{0.45\linewidth}
        \centering
        \includegraphics[width=\linewidth]{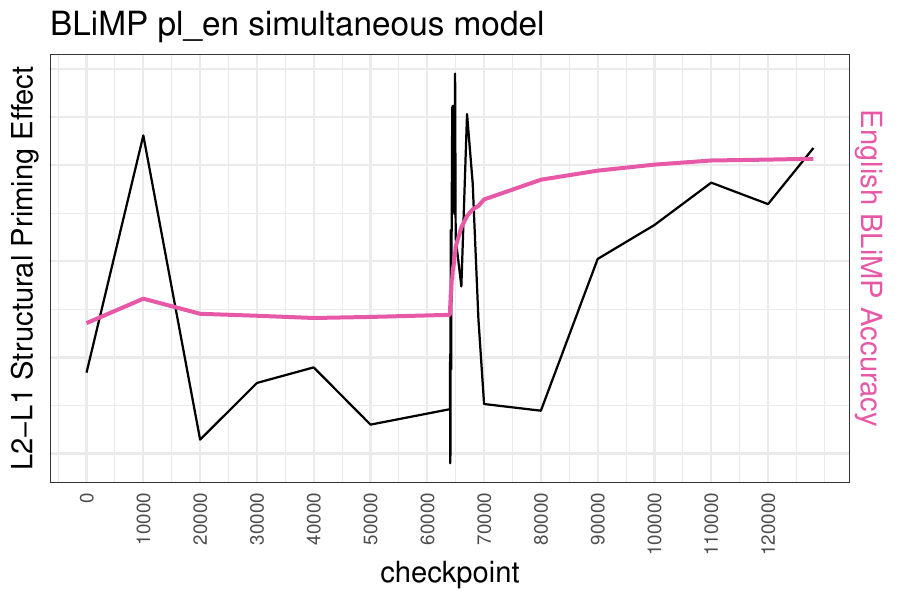} 
        \vspace{0.5em}
        \includegraphics[width=\linewidth]{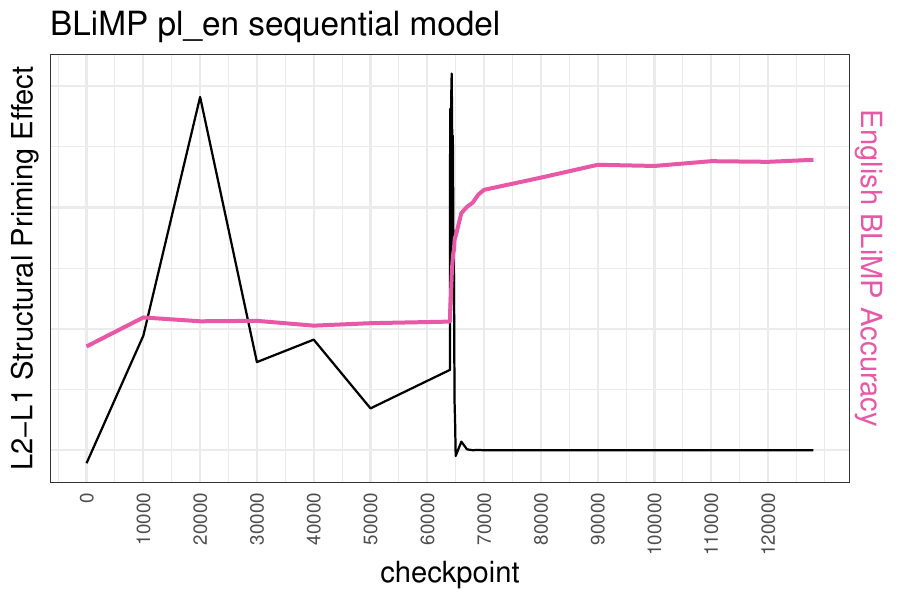} 
        \vspace{0.5em}
    \end{minipage}
    \caption{Structural priming effect (black), plotted as the difference between match and mismatch conditions, and English BLiMP accuracy (pink) over the course of model training. Y-axes have been re-scaled for easier comparison.}
    \label{fig:fleischer_blimp_full}
\end{figure*}

Figure \ref{fig:fleischer_blimp_full} shows the comparison for structural priming effects and BLiMP scores for all Polish-English models. 

\subsection{Kotzochampou (2022)}
\begin{figure*}[h]
    \centering
    \begin{minipage}[b]{0.45\linewidth}
        \centering
        \includegraphics[width=\linewidth]{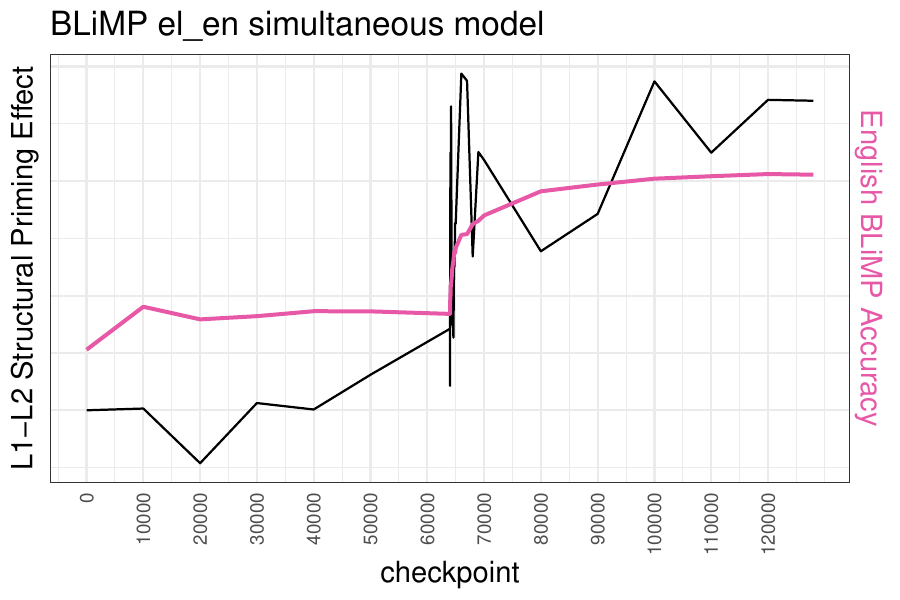} 
        \vspace{0.5em}
        \includegraphics[width=\linewidth]{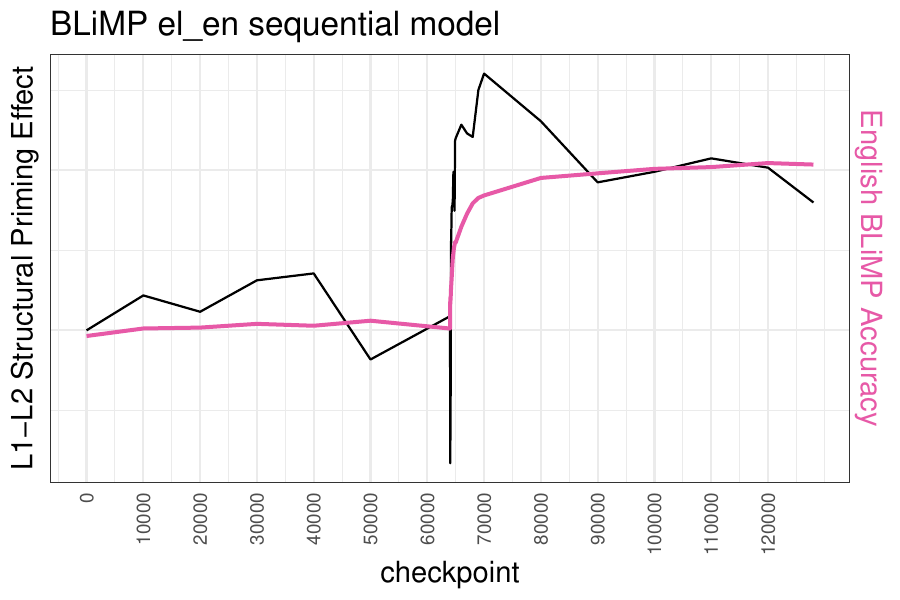} 
        \vspace{0.5em}
    \end{minipage}
    \hfill 
    \begin{minipage}[b]{0.45\linewidth}
        \centering
        \includegraphics[width=\linewidth]{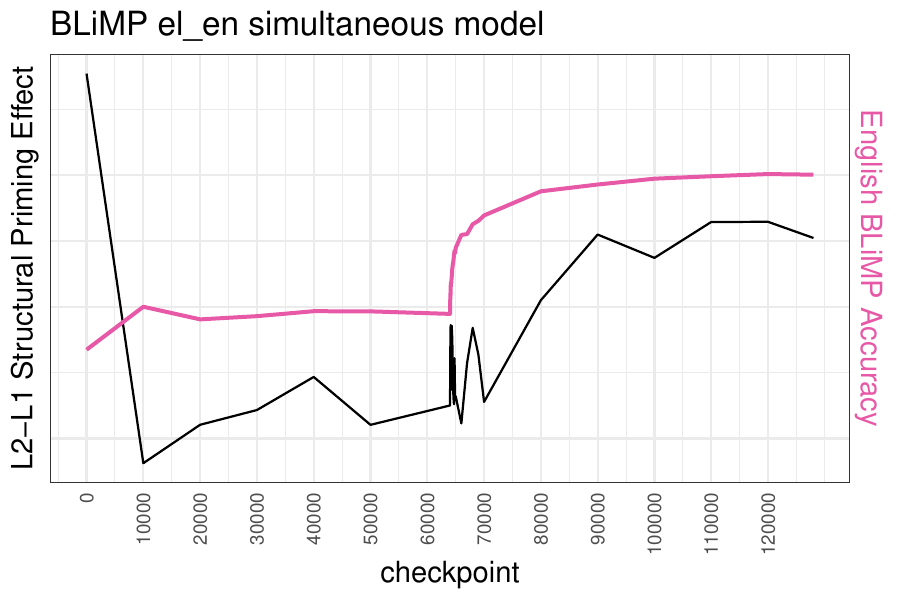} 
        \vspace{0.5em}
        \includegraphics[width=\linewidth]{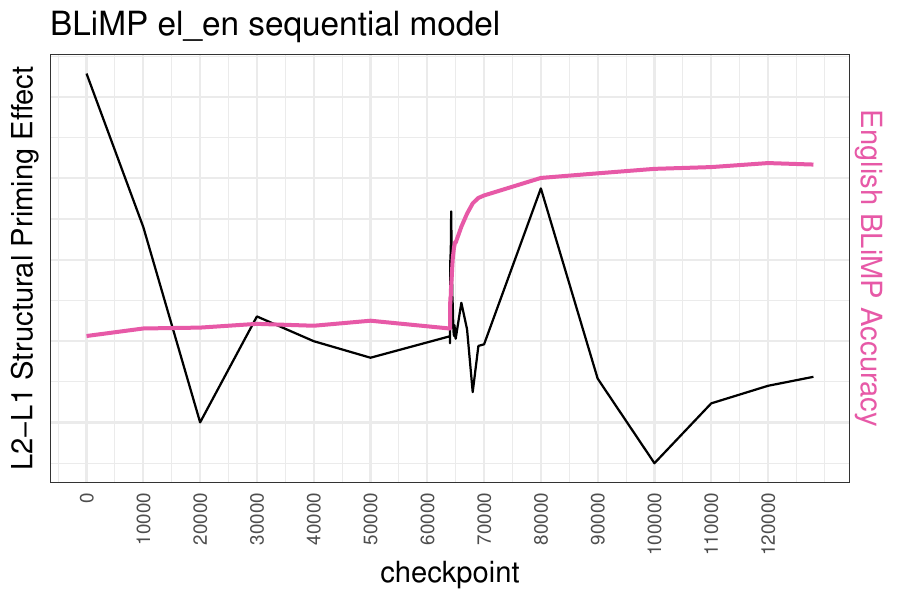} 
        \vspace{0.5em}
    \end{minipage}
    \caption{Structural priming effect (black), plotted as the difference between match and mismatch conditions, and English BLiMP accuracy (pink) over the course of model training. Y-axes have been re-scaled for easier comparison.}
    \label{fig:kotzochampou_blimp_full}
\end{figure*}

Figure \ref{fig:kotzochampou_blimp_full} shows the comparison for structural priming effects and BLiMP scores for all Greek-English models. 

\section{Supplementary BLiMP Analysis} \label{sec:supplementary_blimp}
\renewcommand{\thetable}{F.\arabic{table}}
\renewcommand{\thefigure}{F.\arabic{figure}}

\begin{figure*}[h!]
    \centering
    \includegraphics[width=0.75\linewidth]{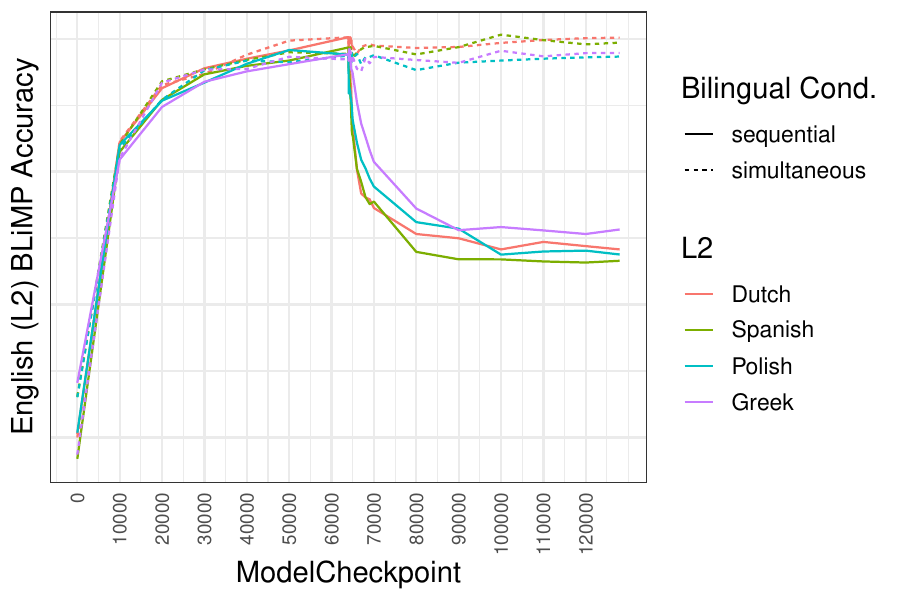}
    \caption{English L1 models in both the sequential (solid lines) and simultaneous (dotted lines) conditions. BLiMP accuracy is plotted over the course of training. } 
    \label{fig:blimp_eng_l1}
\end{figure*}

For models where English is the L1, we see differences in BLiMP scores over the course of training according to the bilingual conditions (Figure \ref{fig:blimp_eng_l1}). In the simultaneous bilingual condition, there is a small dip in BLiMP score after exposure to L2, but then the scores rise again and stay at ceiling. In the sequential bilingual condition, BLiMP scores fall rapidly after exposure to L2. At about step 80000, performance plateaus. The performance never returns to the level of the model at checkpoint 0, but BLiMP score at the final checkpoint is worse than at checkpoint 10000 for all models. This further supports the observation that the models in the sequential bilingual condition experience catastrophic forgetting. It is even more noteworthy, therefore that the models exhibit structural priming effects during the period where L1 mean surprisal rises and BLiMP scores fall. 

Comparing BLiMP performance for the models in the simultaneous condition, we observe a difference in final checkpoint performance. Dutch models have the best performance, followed by Spanish. Greek and Polish again show the worst performance. These results demonstrate differential crosslingual transfer benefits. The language that is the most similar to English (Dutch) leads to the highest BLiMP scores, followed by Spanish, which is also very similar to English. Polish and Greek are the most different from English and show the least benefit from crosslingual transfer. This is also consistent with previously demonstrated effects of linguistic similarity \citep{chang-etal-2024-multilinguality}.

\end{document}